\documentclass[11pt,a4paper]{article}
\usepackage{times,latexsym}
\usepackage{url}
\usepackage[T1]{fontenc}
\usepackage{listings}
\usepackage[T1]{fontenc}
\usepackage[utf8]{inputenc}
\usepackage{latexsym}
\usepackage{amsfonts}
\usepackage{amsmath}

\usepackage[acceptedWithA]{tacl2021v1}

\usepackage{float}
\usepackage{microtype}
\usepackage{times}
\usepackage{latexsym}
\usepackage{caption}
\usepackage{graphicx}
\usepackage{subfigure}
\usepackage{makecell}
\usepackage{booktabs}
\usepackage{multirow}
\usepackage{amsmath}
\usepackage{subfigure}
\usepackage{booktabs}
\usepackage{makecell}
\usepackage{arydshln}
\usepackage[utf8]{inputenc}
\usepackage[T1]{fontenc}
\usepackage{hyperref}
\usepackage{makecell}
\usepackage{xcolor,colortbl}
\usepackage{xcolor}

\newcommand{\griit}[1]{\textcolor{black}{#1}}
\newcommand{\liang}[1]{\textcolor{black}{#1}}
\usepackage{xspace,mfirstuc,tabulary}

\newif\iftaclinstructions
\taclinstructionsfalse % AUTHORS: do NOT set this to true
\iftaclinstructions
\renewcommand{\confidential}{}
\renewcommand{\anonsubtext}{(No author info supplied here, for consistency with
TACL-submission anonymization requirements)}
\newcommand{\instr}
\fi

\iftaclpubformat % this "if" is set by the choice of options

\else

\fi

\title{Unifying Structured Data as Graph for Data-to-Text Pre-Training}

\author{
  Shujie Li $^{1}$\footnotemark[1]
\quad Liang Li $^{3}$
\quad Ruiying Geng $^{2}$ 
\quad Min Yang $^{1}$\footnotemark[2]
\quad Binhua Li $^{2}$ 
\\{\bf
\quad Guanghu Yuan $^{1}$
\quad Wanwei He $^{1}$
\quad Shao Yuan $^{2}$
\quad Can Ma $^{3}$
\quad Fei Huang $^{2}$
\quad Yongbin Li$^{2}$\footnotemark[2]}
\\
$^1$ Shenzhen Institute of Advanced Technology, Chinese Academy of Sciences \\
$^2$ DAMO Academy, Alibaba Group \\
$^3$ Institute of Information Engineering, Chinese Academy of Sciences \\
\tt{\{sj.li1, min.yang\}}@siat.ac.cn, \tt{shuide.lyb}@alibaba-inc.com
}
\date{}

\begin{document}

\maketitle

\renewcommand{\thefootnote}{\fnsymbol{footnote}}
\footnotetext[1]{Work done during an intern at Alibaba DAMO Academy.}
\footnotetext[2]{Corresponding author.}
\renewcommand{\thefootnote}{\arabic{footnote}}

\begin{abstract}
Data-to-text (D2T) generation aims to transform structured data into natural language text. Data-to-text pre-training has proved to be powerful in enhancing D2T generation and yields impressive performances. However, previous pre-training methods either oversimplified structured data into a sequence without considering input structures or designed training objectives tailored for a specific data structure (e.g., table or knowledge graph). In this paper, we unify different types of structured data (i.e., table, key-value data, knowledge graph) into the graph format and cast different data-to-text generation tasks as graph-to-text generation. To effectively exploit the structural information of the input graph, we 
propose a structure-enhanced pre-training method for D2T generation by designing a structure-enhanced Transformer. Concretely, we devise a position matrix for the Transformer, encoding relative positional information of connected nodes in the input graph. In addition, we propose a new attention matrix to incorporate graph structures into the original Transformer by taking the available explicit connectivity structure into account. Extensive experiments on six benchmark datasets show the effectiveness of our model. Our source codes are available at \url{https://github.com/AlibabaResearch/DAMO-ConvAI/tree/main/unid2t}.

\end{abstract}

\section{Introduction}
Data-to-text (D2T) generation, which aims to generate a target natural language text conditioned on source structured data, has attracted noticeable attention due to its wide applications such as journalism \cite{,rebuffel2020hierarchical}, medical diagnosis \cite{nishino2020reinforcement}, financial and weather reports \cite{liang2009learning}, and sports broadcasting \cite{chen2008learning}. The input structured data can include tables of records, simulations of physical systems, spreadsheets, knowledge graphs, and so on. Transforming structured data into textual data can facilitate a wide range of users to understand and use the structured data, which is needed in many real-life scenarios.  

Recently, large-scale pre-trained models have proved to be powerful in D2T generation and yield impressive performances~\cite{kale-rastogi-2020-text,xing2021structure,liu2022plog}, which benefit from the rich knowledge contained in large-scale pre-training corpora.  \citet{xing2021structure} proposed a structure-aware table-to-text pre-training model, which devised three self-supervised training objectives %(i.e., masked table language model, adjacent cell prediction, context reconstruction) 
tailored for modeling tables and their contexts. \citet{ke2021jointgt} adopted a structure-aware semantic aggregation module to model the structure of an input graph at each Transformer layer, and explicitly learned graph-text alignments instead of directly fine-tuning text-to-text pre-trained models on graph-to-text corpora.
%As revealed by \citet{ribeiro2020investigating}, the approaches based on large-scale pre-trained language models (PLMs) could achieve excellent graph-to-text performance even if the input graph was simplified to a bag of nodes and edges without any information about connectivity.

%Despite the significant progress of previous studies
Although significant progress has been made in this field, there are still several technical challenges with existing data-to-text pre-training methods. Most prior studies made a cumbersome design tailored for a specific data structure such as tables \cite{liu2022plog} or knowledge graphs \cite{li2022graph}, which could not effectively deal with diverse structured data in a unified framework. \citet{kale-rastogi-2020-text} was the first work that studied the ``pre-train and fine-tune'' strategy on several benchmarks spanning task-oriented dialogue, table-to-text, and graph-to-text. However, it oversimplified the input structured data into a flat string %(linearization) 
and adopted an original Transformer without capturing the structural information of source structured data. 

%data-to-text generation is one of the major tasks in the natural language generation task paralleled with Text-to-Text generation, both of which aim to produce natural language texts that meet specific communication goals \cite{dong2022survey}. Structured data (such as tables, triples, and key-value pairs) are often used to represent knowledge or abbreviate results, but transcribing it into text can ease people`s understanding and usage, which is needed in many life scenarios, such as journalism, medical diagnosis, financial and weather reports, and sports broadcasting \cite{rebuffel2020hierarchical}.

In this paper, we \textbf{uni}fy the structured data into the graph format for \textbf{d}ata-\textbf{t}o-\textbf{t}ext pre-training (denoted as \textbf{UniD2T}). We convert diverse types of structured data into a unified graph format, keeping the structural information of the structured data. We treat the items in the structured data as a set of nodes and connect the nodes according to the connectivity of the structured data. 
% In particular, for tables,.... ; for knowledge graphs, ....; for key-value pairs, .... \yangmin{Pls briefly describe how to get edges for these three types of structured data.} 
In this way, we can cast various data-to-text tasks as the graph-to-text generation task. 

To effectively encode the graph structure, we propose a structure-enhanced pre-training model, which can be applied to various downstream data-to-text generation tasks. Our proposed data-to-text pre-training model is built upon the T5 model \cite{2020t5}. Since the T5 model is a text-to-text transfer Transformer framework and cannot effectively encode the graph structure, we propose a structure-enhanced Transformer to encode the structural information. 
%, which generally captures interaction information between any node pair via a single self-attention layer. The graph structure cannot be effectively encoded by the T5 model.
Concretely, we propose an explicit position matrix for the Transformer, encoding the relative positional information of connected nodes in the input graph. In addition, we build a new attention matrix to replace the attention mask in self-attention of the original Transformer, which encodes graph structures and takes the available explicit connectivity structure into account. 

Our main contributions are three-fold. (1) We unify diverse types of structured data into a graph format and cast all data-to-text tasks as the graph-to-text generation task taking a graph as input and producing a text as output. 
(2) We propose a structure-aware pre-training method for D2T generation based on the T5 model, which incorporates relative positional information and graph structures into the original Transformer via two new position and attention matrices respectively.   
(3) We conduct extensive experiments on six data-to-text benchmarks and achieve substantially better performance than strong baselines. We believe that the release of our unified data-to-text pre-training model would push forward the research in this area.

\section{Related Works}
\subsection{Data-to-Text Generation}
Data-to-text (D2T) generation aims to produce output texts from structured data and has attracted noticeable attention from the natural language processing (NLP) community \cite{DBLP:journals/nle/ReiterD97}. Recently, neural D2T models \cite{song2018graph,zhu2019modeling} have been the mainstream for this task and made impressive progress. The end-to-end neural models generate text directly from structured data by using an encoder-decoder architecture \citep{DBLP:conf/nips/SutskeverVL14}. These works usually focus on improving the encoder structures based on attention mechanisms \cite{koncel2019text,mehta2022improving} or graph neural networks (GNNs) \cite{philipp2021modeling,ribeiro2021investigating,ribeiro2021structural}.  
%They simply rely on representation learning to improve the generation. 
For example,  \citet{wang2020amr} proposed a graph-to-sequence model using a pairwise interaction function to obtain semantic relations between concepts. \citet{puduppully2022data} suggested a neural architecture that incorporated a planning module to manage high-level information in a logical and meaningful manner. \citet{DBLP:conf/aaai/LiuWSCS18} proposed a structure-aware sequence-to-sequence architecture, which incorporated the filed information as additional input to the table encoder. 
%Some works design hierarchical encoder which model structural representation from the original data \cite{DBLP:conf/emnlp/GongFQL19}. 
\citet{song2018graph} introduced graph recurrent networks (GRNs) to encode the AMR nodes directly. Subsequently, \citet{shi2020g2t} proposed GNNs as the structural encoder, which updated the representations of nodes based on their immediate neighbors. 
To integrate both local and non-local features and learn a better structural representation of a graph, \citet{guo2019densely} introduced dense connection and allowed deep GCNs. 
Different from the local information aggregation scheme, \citet{cai2020graph} proposed a graph transformer that used explicit relation encoding and allowed direct communication between two distant nodes.

\subsection{Data-to-Text Pre-training Models}
Recently, we have witnessed the remarkable success of pre-training methods in a wide range of NLP tasks \cite{kenton2019bert,radford2018improving,lan2019albert,bi2020palm}.
Most pre-training models are initially designed to text-to-text generation, lacking the ability to encode structural information. Recently, there exist some pre-training models designed for data-to-text tasks \cite{chen2020kgpt,agarwal2021knowledge,ke2021jointgt,bai2022graph}. For example, KGPT~\cite{chen2020kgpt} proposed a distantly supervised learning method to exploit large-scale unlabeled web text for data-to-text pre-training.
%STTP~\cite{xing2021structure} introduced three self-supervised tasks to train a structure-aware table-to-text pre-trained model.
%is a struct-aware pretraining model for table-to-text generation tasks. It is trained with structured input table and their contexts. 
However, these pre-training models consider only one specific data structure and cannot be applied to diverse downstream data-to-text tasks. Although \citep{tang2022mvp} proposed a multi-task supervised pre-training model (MVP) for a series of data-to-text generation tasks, it %oversimplified the input structured data into a flat string and 
utilized the original Transformer to encode the linearized input data without considering the graph structures. 
UniLM~\cite{UniLM} was a pre-trained universal language model, which incorporated modified self-attention masks to facilitate bidirectional encoding or unidirectional decoding. While UniLM offers the flexibility of bidirectional encoding, its encoding attention mask is designed primarily for processing unstructured text, thereby restricting its ability to capture the structural characteristics of input graphs.
%Consequently, UniLM may not be fully equipped to effectively model and represent the inherent structure of input graph data.}
%\liang{UniLM~\cite{UniLM} a pre-trained language model that modifies self-attention masks to enable bidirectional encoding or unidirectional decoding. However, the encoding attention mask in UniLM is bidirectional and primarily suited for encoding unstructured text, which limits its effectiveness in capturing the structural aspects of input graphs. Consequently, it may not adequately model the input graph's inherent structure.}

Different from previous works, we propose a unified pre-training model that casts all data-to-text tasks as the graph-to-text generation task. In addition, we incorporate graph structures into the original Transformer via two new position and attention matrices to effectively model the structured input data.

\section{Pre-training Data Construction}
Previous data-to-text pre-training datasets are usually tailored to specific structured data. In this paper, we collect eight data-to-text datasets from previous works and aggregate these datasets into a large corpus for pre-training our model. The statistics of pre-training data are provided in Table~\ref{dataset_statistic}. 

\begin{table}[t!]
\centering
\resizebox{0.8\columnwidth}{!}{
\begin{tabular}{@{}lcc@{}}
\toprule

    \textbf{Statistics}          & \textbf{\textsc{PreData}} & \textbf{\textsc{DownData}}\\ 
    \midrule
\# Datasets                   & 2 & 6      \\
\# Instances                  & 4,951,267 & 2,240,927  \\
Avg. input tokens      & 84.1 & 63.7  \\
Avg. target tokens      & 90.8 & 100.9  \\
Avg. Nodes              & 17.8 & 19.4  \\
Avg. Edges              & 112.3 & 103.1  \\
\bottomrule
\end{tabular}
}
\caption{\label{dataset_statistic}Statistics of our pre-training data.}
\end{table}

\subsection{Existing Pre-training Datasets (\textsc{PreData})}
We first collect the table-text dataset \textsc{TaPas} \cite{herzig2020tapas} and the graph-text dataset KGTEXT \cite{chen2020kgpt}, which were originally designed for table-to-text and graph-to-text pre-training respectively. \textsc{TaPas} contains 6.2M tables from Wikipedia, while KGTEXT consists of 1.8M hyperlinked senftences from Wikipedia
with the corresponding knowledge subgraphs from WikiData. We further devise a rule-based data-cleaning strategy to guarantee data quality. Finally, we obtain 4.9M data-text pairs (called \textsc{PreData}). 
% The data cleaning process can be found in Appendix~\ref{data_clean}.
% \yangmin{why Table 1 shows that there are 4,951,267 instances rather than 3.2 instances?}

\subsection{Existing Downstream Datasets (\textsc{DownData})}
We also collect the training sets from six data-to-text datasets, including WebNLG \cite{gardent2017creating}, DART \cite{nan2020DART}, ToTTo \cite{parikh2020ToTTo}, WikiBio \cite{lebret2016neural}, WikiTableT \cite{chen2021WikiTableT}, and CoSQL \cite{yu-etal-2019-CoSQL}. These datasets were designed for downstream data-to-text generation tasks. Concretely, WebNLG and DART are graph-to-text datasets; WikiBio and WikiTableT contain key-value pairs; ToTTo and CoSQL are table-based datasets. In total, there are about 2.2M instances (\textsc{DownData}).  \griit{Notably, the test sets utilized for downstream tasks are expressly omitted from the pre-training data, assuring the integrity of our experimental results by eliminating any potential data leakage.}

\begin{figure}
\centering
\includegraphics[height = 1.1\columnwidth ,width =\columnwidth]{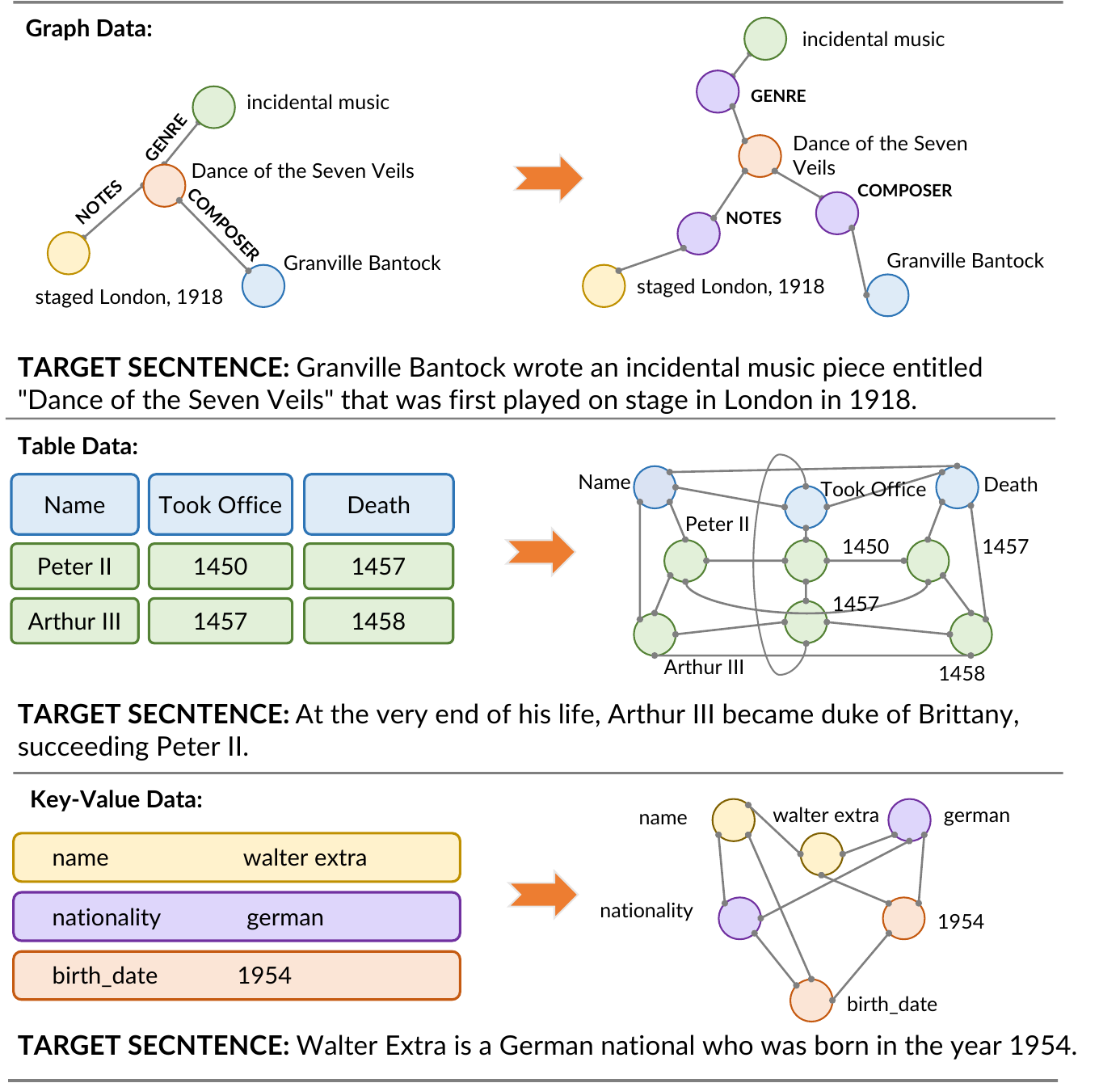}
\caption{\label{fig:unif_process} Unify data in three formats into one graph structure.}

\end{figure}

\subsection{Unifying Structured Data}
\label{sec:unifying_structured_data}
As illustrated in Figure \ref{fig:unif_process}, we unify different structured data (knowledge graph, table, key-value pairs) into %a graph format.
\liang{an unlabeled and connected graph $\mathcal{G} = (\mathcal{V}, \mathcal{E} )$ which consists of a set of nodes $v \in \mathcal{V}$ and unlabeled edges $(v_i, v_j) \in \mathcal{V}$.
Next, we elucidate the process of transforming the three distinct types of data (i.e., knowledge graphs, tables, and key-value pairs) into a unified graph $\mathcal{G}$.
}

 %A knowledge graph is generally stored using the structure of RDF triples (head entity $v_s$, relation $r$, tail entity $v_t$). 
\liang{ (1)
%A knowledge graph $\mathcal{G}^0$ generally consists of a set of labeled edges.
%Each labeled edge is a triple $(v_s, r, v_t)$, where $v_s$ is the head node, $v_t$ is the tail node, and $r$ is the relation between these two adjacent nodes.
On the left side of Figure \ref{fig:unif_process}'s Graph Data section, a knowledge graph can be formally expressed as $\mathcal{G}_0 = (\mathcal{V}_0, \mathcal{E}_0, \mathcal{R}_0)$, where nodes are denoted by $v \in \mathcal{V}_0$, and labeled edges are represented as $(v_s, r, v_t) \in \mathcal{E}_0$, with $r \in \mathcal{R}_0$ signifying the relation type. To more effectively model the relationships between nodes within the knowledge graph $\mathcal{G}_0$ without modifying the underlying model architecture, we transform it into its equivalent Levi graph, as shown on the right side of the Graph Data section in Figure \ref{fig:unif_process}, following similar methodologies as in prior studies~\citep{ribeiro2021structural,li2022graph}.
 A Levi graph is formally characterized as an unlabeled, connected bipartite graph, denoted as $\mathcal{G} = (\mathcal{V}, \mathcal{E})$. Specifically, each relation in $\mathcal{R}_0$ is treated as a new graph node within $\mathcal{G}$ and amalgamated with all nodes in $\mathcal{V}_0$ to form the comprehensive node set $\mathcal{V}$. Subsequently, each edge $(v_s, r, v_t) \in \mathcal{E}_0$ labeled with a relation type is converted into two unlabeled, undirected edges $(v_s, r), (r, v_t) \in \mathcal{E}$. In addition, for each unlabeled edge, corresponding reverse edges  $(r, v_s)$, $(v_t, r)$ are introduced. For instance, given a labeled edge (\textit{Dance of the Seven Veils}, \textit{GENRE}, \textit{incidental music}), this conversion results in four unlabeled edges (\textit{Dance of the Seven Veils}, \textit{GENRE}), (\textit{GENRE}, \textit{Dance of the Seven Veils}), (\textit{GENRE}, \textit{incidental music}), and (\textit{incidental music}, \textit{Dance of the Seven Veils}), comprising the final Levi graph $\mathcal{G}$.
 % $(v_s, r)$, $(r, v_s)$, $(r, v_t)$, and $(v_t, r)$, comprising the final Levi graph $\mathcal{G}$.
%A knowledge graph can be formulated as $\mathcal{G}_0 = (\mathcal{V}_0, \mathcal{E}_0, \mathcal{R}_0)$ with nodes $v \in  \mathcal{V}_0$ and labeled edges $(v_s,r,v_t) \in \mathcal{E}_0$, where $r \in \mathcal{R}_0$ is the relation type. To better model the relations between nodes in the knowledge graph $\mathcal{G}_0 $ while avoiding modifying the model architecture, we transform it into its equivalent Levi graph based on previous works~\cite{ribeiro2021structural,li2022graph}. As illustrated in the first part of Figure~\ref{fig:unif_process}, a Levi graph is an unlabeled and connected bipartite graph and can be defined as $\mathcal{G} = (\mathcal{V}, \mathcal{E})$. Specifically, we regard all relations in $\mathcal{R}_0$ as new graph nodes in $\mathcal{G}$ and combine them with all nodes in $\mathcal{V}_0$ to build the node set $\mathcal{V}$. And then, each labeled edge $(v_s, r, v_t) \in \mathcal{E}_0$ is converted into two unlabeled edges $(v_s, r), (r, v_t) \in \mathcal{E}$. In addition, we add a reverse edge for each unlabeled edge. For example, given a labeled edge $(v_s, r, v_t)$, it is converted to four unlabeled edges, $(v_s, r)$, $(r, v_s)$, $(r, v_t)$, and $(v_t, r)$, in the final Levi graph $\mathcal{G}$.
}
%We treat both entities and relations as nodes in our new graph, and there is an edge between each entity and relation within a triple. 
%For example, given a RDF triple ($s$, $p$, $o$), we have two bidirectional edges: $s \Leftrightarrow p$, $p \Leftrightarrow o$.
%As shown in Figure \ref{fig:unif_process}, we will introduce the process of building unified structure data as nodes and edges according to different data types. For graph data, the original data is stored using the structure of RDF triples ($s$, $p$, $o$). We regard all of them as nodes, and different nodes with the same content have the same id, and establish a bidirectional connection for each RDF according to the relationship of triplets: $s \Leftrightarrow p$, $p \Leftrightarrow o$.

\liang{(2) In the Table Data section of Figure \ref{fig:unif_process}, situated on the left side, Tabular data is conventionally structured with numerous cells organized based on their respective roles and interrelations. A table can be formally represented as $\mathcal{T} = {v_{i,j}|i\in [1, N], j\in [1, M]}$, where $v_{i,j}$ denotes a table cell, and $N$ and $M$ represent the number of rows and columns in the table, respectively. Inspired by recent studies~\cite{wang2022robust,li-etal-2023-cats}, we employ a heuristic rule to transform the tabular data into a unified graph $\mathcal{G}$ by introducing unlabeled edges between cells based on their roles and relationships. This structural transformation serves to maintain the invariance of the table content and proficiently articulate the relationships among cells in the table. More precisely, all cells within $\mathcal{T}$ are considered as graph nodes in $\mathcal{G}$, denoted as $\mathcal{V} = \mathcal{T}$. Furthermore, we establish the set of unlabeled edges $\mathcal{E}$ in accordance with two guiding principles. First, for any two cells $v_{i,j}$ and $v_{i,z}$ situated within the same row, we introduce a forward edge $(v_{i,j}, v_{i,z})$ along with a corresponding reverse edge $(v_{i,z}, v_{i,j})$ into $\mathcal{E}$. Second, for any two cells $v_{i,j}$ and $v_{i,z}$ located in the same column, we append a forward edge $(v_{i,j}, v_{z,j})$ and its corresponding reverse edge $(v_{z,j}, v_{i,j})$ to $\mathcal{E}$. For instance, contemplating the right Table Data section in Figure \ref{fig:unif_process}, the cell ``Arthur III'' is linked not only to cells ``1457'' and ``1458'' in the same row but also to cells ``Name'' and ``Peter II'' in the same column. This intentional configuration is based on empirical observations and insights gained from data analysis. However, we acknowledge that there exists room for further exploration and experimentation concerning diverse node connectivity settings in future research. Given that the ToTTo dataset exclusively generates text for highlighted data, only the highlighted cells are considered as nodes.}

(3) \griit{For Key-Value data in Figure \ref{fig:unif_process}, both key and value are regarded as nodes within $\mathcal{V}$. In addition to the requisite connection edges linking each (key, value) pair (e.g., the connection between the key \textit{name} and the value \textit{walter extra}), we extend our connectivity framework to include connections among keys themselves (e.g., the connection between \textit{nationality} and \textit{birth\_date}) and value themselves(e.g., the connection between \textit{walter extra} and \textit{gernman}),  drawing inspiration from the graph construction methodology commonly employed in table data analysis. In line with tabular data, we introduce both forward and reverse edges for any connected nodes within $\mathcal{V}$.} 

\begin{table}[]
\setlength{\tabcolsep}{3pt}
\resizebox{\columnwidth}{!}{
\begin{tabular}{@{}cll@{}}
\toprule
Type           & Dataset & Prefix-S                 \\ \midrule
\multirow{2}{*}{Table} & ToTTo      & \begin{tabular}[c]{@{}l@{}}The table page title is: A,\\ The table section title is: B\end{tabular}                   \\ \cmidrule{2-3} 
                                   & CoSQL   & select A from B where C  \\ \midrule
\multirow{2}{*}{Graph}                      & DART    & The source is: A         \\ \cmidrule{2-3} 
                               & WebNLG     & \begin{tabular}[c]{@{}l@{}}The category of the entities is: A,\\ The number of RDF triples is: B\end{tabular} \\ \midrule
\multirow{2}{*}{Key-Value} & WikiBio & The article title is:  A \\ \cmidrule{2-3} 
                               & WikiTableT & \begin{tabular}[c]{@{}l@{}}the document title is: A,\\ the section title is: B\end{tabular}                           \\ 
\bottomrule
\end{tabular}
}
\caption{\label{tab:prefix} The data-specific prefixes that are tailored for different types of data. Here, A, B and C can be replaced by the content of specific samples.}
\end{table}

\griit{To ensure clarity and context in the generated text, we introduce two specific prefixes before the actual input data: (1) A data-independent prefix which universally states ``describe the following data.'' (2) A data-specific prefix, tailored according to the nature and structure of the data at hand. We provide the data-specific prefixes for the three data structures in Table \ref{tab:prefix}.}
%\yangmin{give the prefix of three types of data}. 
%Specifically, one is to use other information in the datasets as a special prefix, and the other is to add ``describe the following data: '' to all datasets as a unique prefix. Next, add special tags [Prefix] and [Node] before prefix and node. 
For example, the triple ``$Jens\_Hartel$ | $club$ |$ Berliner\_AK\_07$'' from the DART dataset will add the common prefix and its special prefix to form an input ``\textit{[Prefix] describe the following data: [Prefix] The category of the DBpedia entities is : SportsTeam. [Node] Jens\_Hartel [Node] club [Node] Berliner \_AK\_07}''. 
\begin{figure}[!t]
    \centering
    \includegraphics[width=1.0\columnwidth]{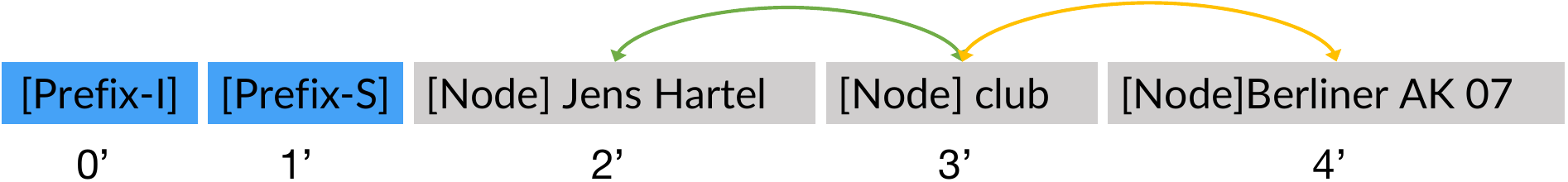}
    \caption{Simplified version of model input and connections between nodes.}
    \label{fig:connection.pdf}
\end{figure}
% \verb|$[Prefix] describe the following data: [Prefix] The category of the DBpedia entities is: SportsTeam. [Node] Jens\_Hartel [Node] club [Node] Berliner \_AK\_07$|. 
%In order to facilitate the understanding of the construction of the graph, 
We simplify the data-independent and data-specific prefixes to ``[\textit{Prefix-I}]'' and ``[\textit{Prefix-S}]'',  respectively. The final  input sequence with connectivity information is shown in Figure \ref{fig:connection.pdf}.
%pretrain  finetune
\section{Methodology}
\label{section:USDG}
% \subsection{Input Construction}
\subsection{Problem Definition}
%\yangmin{Pls modify this section to introduce the input representation. I believe the following content should be merged with Section 3.3.}
We convert different structured data into a graph format and cast all data-to-text tasks as the graph-to-text (G2T) generation task. Formally, the G2T model takes a graph $\mathcal{G}=(\mathcal{V}, \mathcal{E})$ as input and produces a text $Y=\{y_1, \ldots, y_n\}$ as output, where $\mathcal{V}$ represents the entity set, $\mathcal{E}$ represents the relations between entities, and $n$ is the length of the output text. Following previous studies \cite{ribeiro2020investigating}, we convert the graph $\mathcal{G}$ into an input sequence $\mathcal{G}_{\rm linear}=\{x_1, \ldots, x_m\}$ consisting of $m$ tokens. 

% After obtaining unified structured data, we will add two hard prefixes before the input data, including a data-independent prefix and a data-specific prefix. Specifically, the data-independent prefix is ``\textit{describe the following data:}'', which will be added to all the datasets. The data-specific prefix is tailored for each data structure. We provide the data-specific prefixes for the three data structures in Table \ref{tab:prefix}. 

%and the final connection is shown in the figure below:
% in Figure \ref{fig:unif_process}, The example input of graph data is:``\textit{[Prefix] describe the following data: [Prefix] the category of the RDF is:music. [Node] Dance of the Seven Veils [Node] GENRE [Node] incidental music [Node] COMPOSER [Node] Granville Bantock [Node] NOTES [Node] staged London, 1918.} ''

%More input examples and prefix information for other datasets will be placed in the appendix.

% We define a certain position of the two matrices as $\{x, y\}$ = k, where x represents its own node, y represents the rest of the nodes, and {x,y} represents the relationship between x and y.
% For the attention matrix, k = $\{ 0, 1 \}$, 1 means that there is attention, and 0 means that it does not exist.
% For the position matrix, k = $\{ -n, n \}$, n is the sum of the lengths of the largest two nodes corresponding to the edge. When n>0, it means that y is k units to the right of x, and when n<0, it means that y is k units to the left of x.

\subsection{Model Architecture} 

Our model is built upon the pre-trained T5 model given the impressive performance of T5 on text generation tasks. It is noteworthy that our pre-training strategy is model-agnostic and potentially applicable to any Transformer-based backbone networks. The encoder of Transformer is composed of a stack of blocks, each of which contains a self-attention layer followed by a feed-forward network. The decoder has a similar structure to the encoder except that it adopts a standard attention mechanism following a self-attention layer.

\begin{figure}[t!]
\centering
\includegraphics[width=1.0\columnwidth]{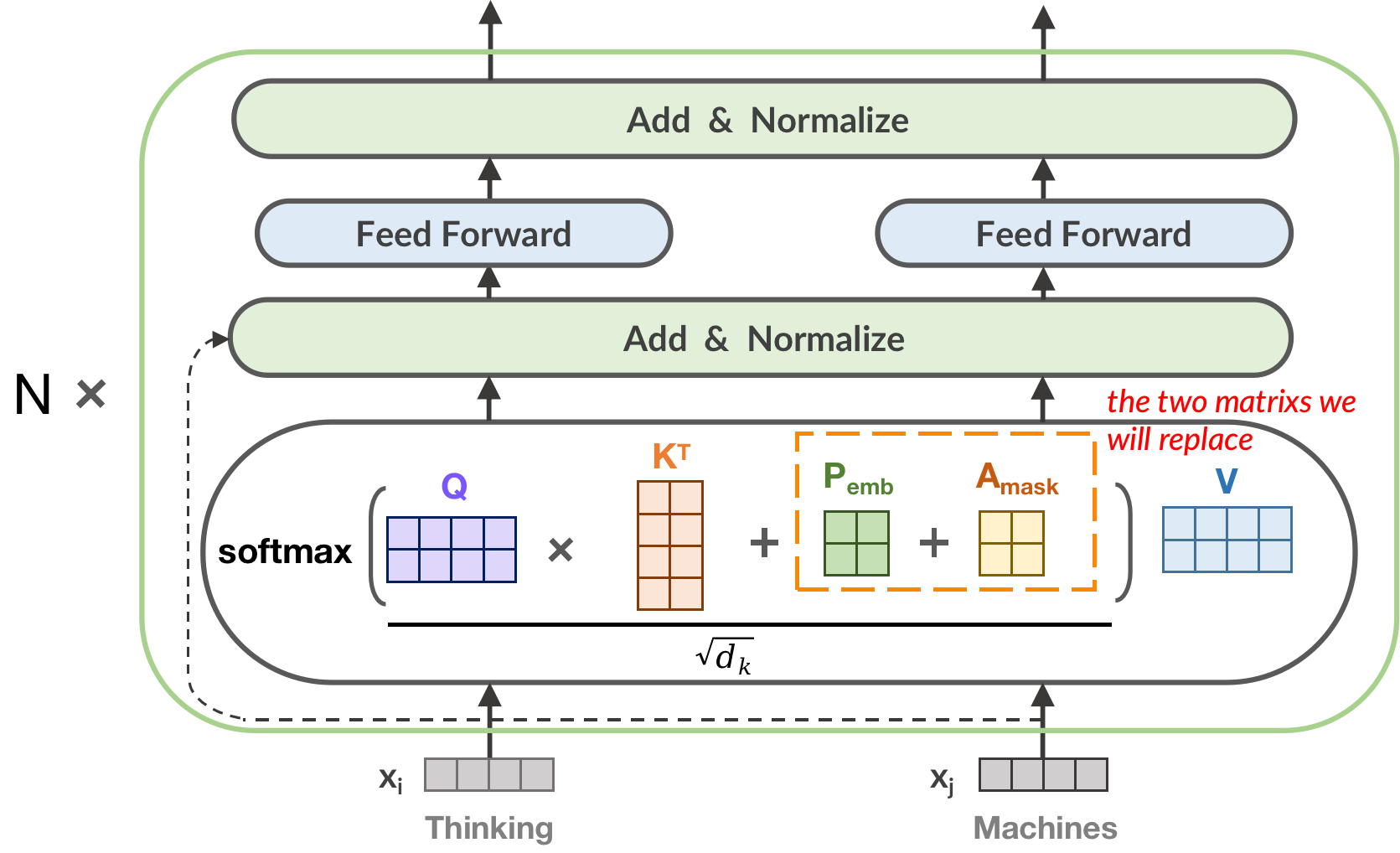}
\caption{Transformer blocks on the T5-encoder side. The  relative position and attention matrices in the self-attention calculation will be replaced by two novel position and attention matrices.}
\label{fig:architecture}
\end{figure}

\griit{\noindent \textbf{Preliminary~} 
% The original self-attention in Transformer is order-independent.
In the case of the T5-encoder, a ``fully-visible'' attention mask is employed, which permits the self-attention mechanism to consider all input entries when generating each output entry. 
In addition, T5 adopts a simplified form of position embeddings, where each embedding is a scalar. 
Formally, as illustrated in Figure~\ref{fig:architecture}, the attention calculation of encoder can be expressed as:
\begin{align}
\mathbf{Q} &= \mathbf{X} \mathbf{W}^Q,  \mathbf{K} = \mathbf{X} \mathbf{W}^K,  \mathbf{V} = \mathbf{X} \mathbf{W}^V \\ 
\label{equ:attention}
\alpha &= \frac{1}{\sqrt{d}} \left( \mathbf{Q} \mathbf{K}^T + \mathbf{P}_{\text{emb}} + \mathbf{A}_{\text{mask}} \right) \\ 
\mathbf{Z} &= \frac{\exp(\alpha) }{\sum \exp(\alpha)} \times \mathbf{V}
\end{align}
% \begin{align}
% Q &= XW^Q \\
% K &= XW^K \\
% V &= XW^V \\
% \alpha_{ij} &= \frac{\exp\left(\frac{QK^T}{\sqrt{d_k}}\right)}{\sum_{j'=1}^{n} \exp\left(\frac{QK^{T}}{\sqrt{d_k}}\right)} \\
% \text{Attention}(Q, K, V) &= \alpha \times V
% \end{align}
where $\mathbf{X}$ is the input sequence. $\mathbf{W}^Q\in \mathbb{R}^{d \times d_Q}$, $\mathbf{W}^K\in \mathbb{R}^{d \times d_K}$ and $\mathbf{W}^V\in \mathbb{R}^{d\times d_V}$ are learnable project parameters. $\alpha$ is the attention weight between the query vector $\mathbf{Q}$ and the key vector $\mathbf{K}$. $d$ is the dimensionality of the hidden representations. $\mathbf{Z}$ is the output of the attention module. $\mathbf{P}_{\text{emb}}$ is position embedding and $\mathbf{A}_{\text{mask}}$ is attention mask. 
}

\griit{
% These embeddings fail to capture the relative positional information of nodes in the graph and attention mask cannot capture the structural  relations between nodes in the graph,
The original attention mechanism is designed to process unstructured natural language texts proves inadequate in effectively capturing the inherent structures within graphs. To better process our structured graph data, we replace the position embeddings $\mathbf{P}_{\text{emb}}$ and attention mask $\mathbf{A}_{\text{mask}}$ in the Equation (2) with two new position and attention matrices respectively, ensuring their awareness of the underlying graph structures. Next, we will elaborate on the processes of constructing the position and attention matrices. }  %As illustrated in Figure \ref{fig:architecture}, 

\subsection{Structure-enhanced Transformer}
%As illustrated in Figure \ref{fig:architecture}, 
T5 is based on an encoder-decoder Transformer, which does not necessarily capture graph structures. To address this issue, we propose a structure-enhanced Transformer, which is built upon the new position and attention matrices on the T5 encoder side. 
\griit{As illustrated in Figure \ref{fig:architecture},  we use new position embedding and attention mask matrices (denoted as $\mathbf{P}_{\text{emb}}^{\text{new}}$ and $ \mathbf{A}_{\text{mask}}^{\text{new}}$) to replace the $\mathbf{P}_{\text{emb}}$ and $ \mathbf{A}_{\text{mask}}$ in the Equation~\ref{equ:attention}, respectively. }
%The vertical and horizontal axes represent the relationship between itself and other nodes, respectively.
Specifically, we devise a position matrix for the Transformer to encode the relative positional information of connected nodes in the original input graph $\mathcal{G}$. 
%The colored cell in the $i$-th row and $j$-th column indicates the element $j$ that the self-attention mechanism allows $i$ to attend to, and the blank cell indicates that the element $j$ is not allowed to attend. 
In addition, we propose a new attention matrix to replace the attention mask in the self-attention, which takes the available explicit connectivity structure of the input graph into account. 
%The colored cell in the i-th row and column $j$ indicates the relative position of $j$ to $i$, the positive value means that $j$ is on the right of $i$, the negative value means that it is on the left, and the blank cell is set to $+inf$ or $-inf$ to represent the distance between the two is very far.
%Next, we will elaborate on the processes of constructing the position and attention matrices. 

%In order to adapt to the data structure on the encoder side and make it absolutely insensitive to long distances, and reduce the impact of T5’s original training on the serialized input on the data structure, we change the attention mask and position required by the block in the self-attention calculation on the T5 encoder side.

%We use an example to express the establishment process of the matrix more clearly, in which the input is: "[Prefix] SportsTeam [Node] Jens Hartel [Node] club [Node] Berliner AK 07", and the edge connection is established for (Jens Hartel, club), (club, Berliner AK 07).
\begin{figure}[t!]
    \centering
    \includegraphics[width=0.95\columnwidth]{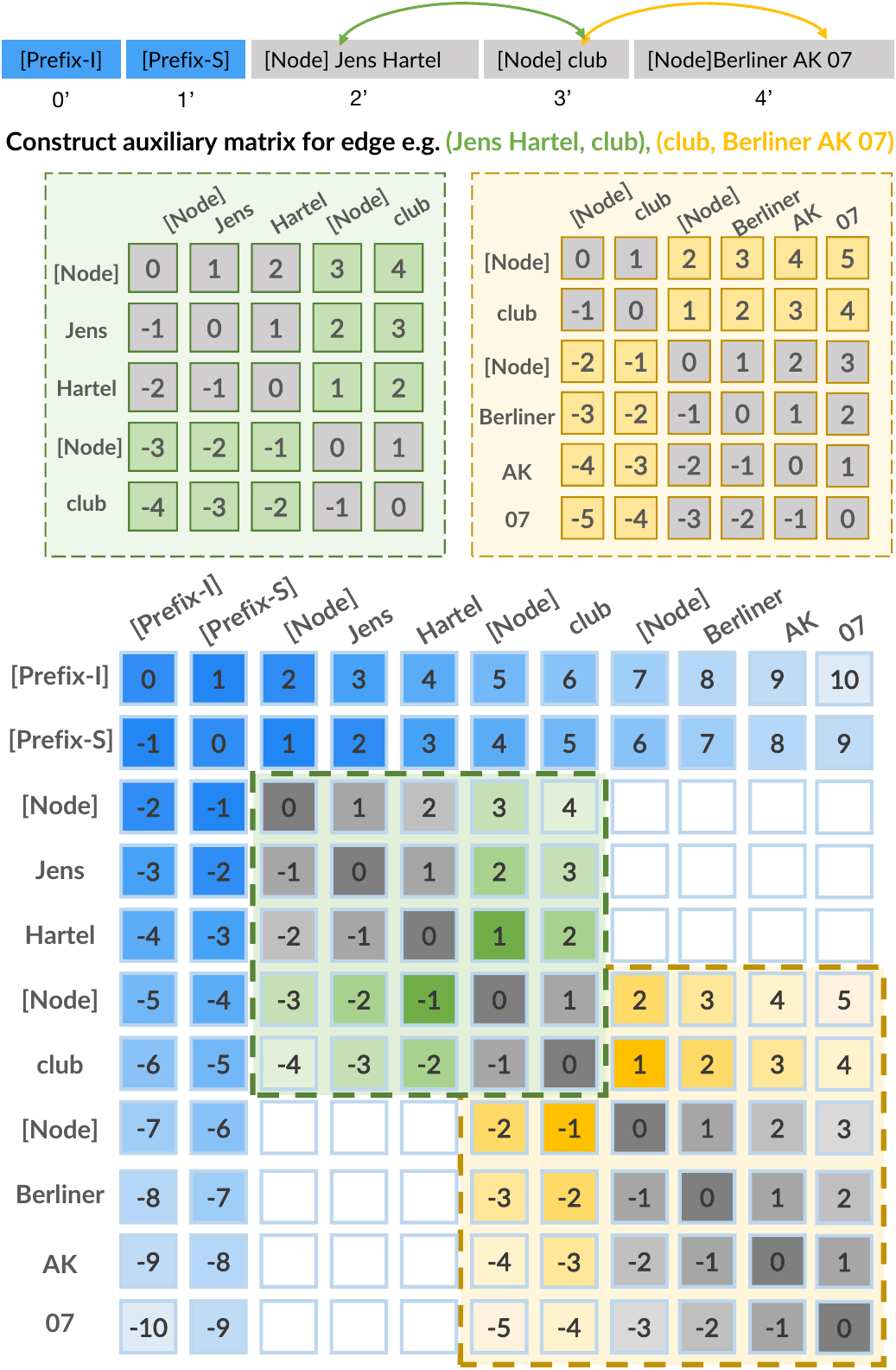}
    \caption{We construct a new position matrix $\mathbf{P}_{\text{emb}}^{\text{new}}$ to replace the original position matrix $\mathbf{P}_{\text{emb}}$ used in Equation (2). We first set an auxiliary matrix for each edge between two nodes, and then copy the content of the auxiliary matrix into the final position matrix. \textbf{\textit{The distances of nodes lacking direct connections will be set to ``$\pm \textbf{inf}$''.}} The lighter the color, the farther the distance is.
    }
    \label{fig:position}
\end{figure}

\subsubsection{Position Matrix Construction}
Integrating relational information about the
graph structure into the Transformer architecture is essential for graph-to-text generation.  Nevertheless, most previous Transformer-based methods \cite{xing2021structure,han2022self_tripple} learned position embeddings automatically, instead of explicitly encoding the structural relationships. For the input graph, we should only consider the relative position between connected nodes but ignore the relative position between irrelevant nodes. To this end, we replace the positional embeddings of the original Transformer with a position matrix that only establishes the relative position between  each relevant node pair (connected items). In this way, we can explicitly capture the relative positions of all relevant nodes precisely.  

Specifically, we first establish an auxiliary position matrix for each pair of connected nodes, similar to the green and yellow boxes in Figure \ref{fig:position}.
\griit{
No matter how physically distant the two relevant nodes may be, the corresponding auxiliary position matrix solely takes into account the relative distance between these two nodes' internal tokens, disregarding the nodes situated between the two target nodes. For example, consider the input nodes ``\textit{[Node] club}'' and ``\textit{[Node] Jens Hartel}'', since ``\textit{club}'' is 3 units to the right of ``\textit{Jens}'', the value of cell [\textit{Jens}, \textit{club}] is 3. Notably, we only compute the relative distance between each connected note pair, while the distances of nodes lacking direct connections will be set to ``$\pm inf$'', signifying an infinite distance between them. For instance, the value assigned to the cell [\textit{Jens}, \textit{Berliner}] is ``$+inf$'' due to the absence of a direct connection between ``\textit{[Node] Jens Hartel}'' and ``\textit{[Node] Berliner AK 07}''. 
% Here, we hope that the prefixes (i.e., [Prefix-I] and [Prefix-S]) within the input can carry global information, thus we make the prefixes attend to all other elements.
%Furthermore, it is essential to emphasize that even in the presence of intervening nodes between nodes ``\textit{[Node] club}'' and ``\textit{[Node] Jens Hartel}'', the value of this position remains unchanged, denoted as 3, signifying a focus solely on the relative position between these two nodes, rather than considering the global context.
}
% \grit{No matter how far the two relevant tokens are actually separated, the corresponding auxiliary position matrix only considers the relative distance between these two tokens, without considering the tokens between the two target tokens. For example, in the input sequence ``\textit{[Prefix] SportsTeam [Node] Jens Hartel [Node] club [Node] Berliner AK 07}'', since ``\textit{club}'' is 3 units to the right of ``\textit{Jens}'', thus the value of cell [\textit{Jens}, \textit{club}] is 3.
% In this way, we can construct an auxiliary position matrix for the connected items (``\textit{Jens Hartel}'', ``\textit{club}''), as demonstrated in Figure \ref{fig:position}.}
After obtaining the auxiliary position matrix for each pair of connected items, we can construct the position matrix for the entire input sequence by copying the cell values from the corresponding auxiliary position matrices. 
\griit{It is noteworthy that we seek to endow the prefixes (denoted as ``[\textit{Prefix-I}]'' and ``[\textit{Prefix-S}]'') embedded within the input with the capacity to encapsulate comprehensive global information. Therefore, we postulate that these prefixes establish direct connections with other nodes within the input. Finally, we replace the positional embeddings $\mathbf{P}_{\text{emb}}$ of original Transformer with the learned position matrix $\mathbf{P}_{\text{emb}}^{\text{new}}$, so as to effectively capture the explicit relative distance between each pair of connected items.}

\begin{figure}[t!]
 \centering
 \includegraphics[width=0.95\columnwidth]{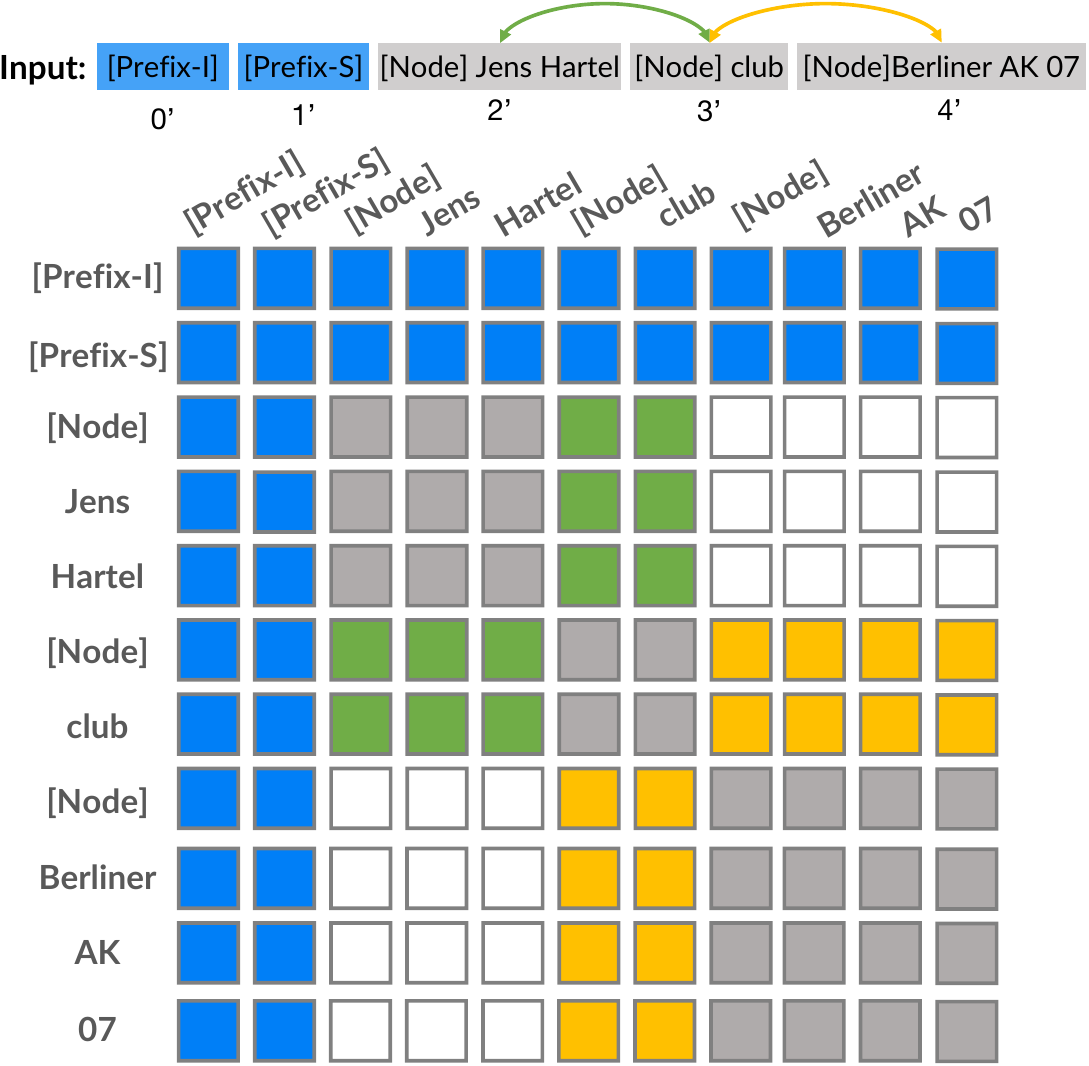}
\caption{We construct a new attention matrix $\mathbf{A}_{\text{mask}}^{\text{new}}$ to replace the attention mask $\mathbf{A}_{\text{mask}}$ used in Equation (2). The attention matrix used to replace the attention mask of self-attention in Transformer. \textbf{\textit{The values of the cells with colors are set to 1, while the values of the cells without colors are set to 0}}. The blue color represents global attention, the gray color represents the self-connection of nodes, and the green and yellow colors represent the two connected edges. }
\label{fig:meta_process}
\end{figure}

\subsubsection{Attention Matrix Construction}
The self-attention in the original Transformer  processes the input sequence by 
 \griit{transforming the input sequence through the substitution of each element with a weighted average. 
 %Conceptually, within a graphical representation, each node ought to be intimately linked solely with its direct connected nodes. 
 Without refining the conventional attention mechanism, the present input data would be perceived as a fully interconnected graph, potentially hindering the optimal extraction of inherent structural information. Given the above reasons, } 
% replacing each element with a weighted average of the rest of the sequence. 
% It generally captures the relationships between any element pairs but cannot explicitly capture the connectivity of each pair of relevant elements. 
% In the graph, each node should be associated exclusively with its directly connected nodes. Without adjustments to the original attention mechanism, the current input data will be treated as a fully connected graph, thus impeding the effective capture of the structural information within the graph.To effectively capture graph structures, 
we construct a relation-aware attention matrix to replace the original attention mask in self-attention. 
Concretely, if two elements have a direct relationship, we set the value of the corresponding cell to 1; otherwise, the value is set to 0. For example, as illustrated in Figure \ref{fig:meta_process}, since the items ``\textit{Jens Hartel}'' and ``\textit{club}'' have direct connection, the values of cells (\textit{Jens}, \textit{club}) and (\textit{Hartel}, \textit{club}) are set to 1; while since ``\textit{Jens Hartel}'' and ``\textit{Berliner Ak 07}'' have no direct connection, the values of the corresponding cells such as (\textit{Jens}, \textit{Berliner}) and (\textit{Jens }, \textit{AK}) are set to 0. Here, we hope that the prefixes (i.e., ``[\textit{Prefix-I}]'' and ``[\textit{Prefix-S}]'') within the input can carry global information, thus we make the prefixes attend to all other elements. \griit{After obtaining the attention matrix (denoted as $\mathbf{A}_{\text{mask}}^{\text{new}}$), we replace the attention matrix $\mathbf{A}_{\text{mask}}$ of self-attention in Equation (2) with our new attention matrix $\mathbf{A}_{\text{mask}}^{\text{new}}$ so as to effectively capture the graph structures as shown in Figure \ref{fig:architecture}.}

%Different from the "full-visible" setting on the T5 encoder side, when the output of the i-th position is generated, we mask the nodes that have nothing to do with it (that is, neither its own node nor a node connected to it by an edge) to prevent $USD$ attends to other unrelated nodes.

%But for prefix, refer to the setting of T5 decoder and ..., we also use fully-visible masking during the prefix portion of the input. Concretely, In Figure 2, the color cell means that it attend to other nodes, so for prompt like prefix and $[Node]$, we hope that it can carry global attention, so it can attend to all other nodes.

%In short, a node will first attend to itself, and then if it has an edge connection with other nodes, it will attend to other nodes.

\subsection{Pre-training Objectives} 
Similar to \cite{andrejczuk2022table}, we first use the publicly available T5 checkpoints provided by \citet{herzig2020tapas} as the initialization. Then, we pre-train our model on our pre-training data. We employ two objectives to pre-train our model in a multi-task learning paradigm, including struct denoising and text generation objectives. In Table~\ref{tab:model-object}, we provide two specific training instances (input and output pairs) for the struct denoising and graph-to-text generation objectives.

%In order to adapt to the data-to-text task, we unify the dataset into the same data representation, and redesign the attention mask and position matrix in T5. The specific design is introduced in the section. Then we randomly divide the data into two pre-training tasks, where the pre-training target is defined as:

\begin{table*}[]
\resizebox{2.1\columnwidth}{!}{
\begin{tabular}{@{}llll@{}}

\toprule
\textbf{Task} &
  \textbf{Inputs} &
  \textbf{Targets} &
   \\ \hline
Struct Denoising &
  \makecell[l]{The category of the DBpedia entities is: $<extra\_id_{0}>$.\\ `Bakewell pudding', `dish variation', `$<extra\_id_{1}>$',\\ `main ingredients', `Ground almond, jam, butter, eggs'} &
  \makecell[l]{$<extra\_id_0>$ Food \\$<extra\_id_1>$ Bakewell tart} &
   \\ \midrule[0.1pt]
Graph-to-Text Generation &
  \makecell[l]{ Describe the following data: The category of the\\ DBpedia entities is : Food. `Bakewell pudding',\\ `dish variation', `Bakewell tart', `main ingredients',\\ `Ground almond, jam, butter, eggs'} &
  \makecell[l]{Bakewell tart is a variation of Bakewell\\
  pudding and  some of  the main ingredients \\ are ground almonds,
  jam, butter and eggs.} &
   \\ \bottomrule
\end{tabular}
}
\caption{\label{tab:model-object}The examples of input-output pairs for struct denoising and graph-to-text generation objectives.}
\end{table*}

\paragraph{Struct Denoising Objective} 
We design a struct denoising strategy for table-like data, following the method used in T5, by training the model to predict a target sequence containing the missing or corrupted tokens in the input graph. 
We apply a noise function to construct a noisy input graph. In particular, the noise function is implemented by masking 15\% of nodes while maintaining related edges in the graph. 
% Inspired by the denoising objective proposed by T5, we propose a struct denoising objective to facilitate representation learning of the input data. In particular, we replace 15\% of nodes in the input with a special \textsc{[mask]} token to construct a perturbed input sequence.
The goal of struct denoising objective is to reconstruct the target output that contains all the dropped-out nodes, delimited by the sentinel token.
% Then, our model is trained to generate the masked sequences. 
This pre-training objective helps the UniD2T model capture relationships between neighboring nodes in the input graph. 
%\yangmin{Why we call this object as struct denoising? I do not see how to use ``struct''.}

%We use 40\% of the unified data set for self-supervised learning and this part of the data does not need to be added to the unified promt. Inspired by Bert's mlm task and in order to adapt to the encoder-decoder model structure, we destroy 15\% of this part of the data set and replace it with a special token tag, then the target is changed to the specific tag and its corresponding content.

\paragraph{Graph-to-Text Generation Objective} 
%We use 60\% of the unified data set for supervised learning.  The input is only linear node and does not deliberately distinguish which data set it comes from. Add a unified promt:``describe the following data:'' before each piece of data, and use the original text as target.

Given the linearized graph $\mathcal{G}_{\rm linear}$ and its explicit connectivity structure $\mathcal{E}$, the graph-to-text generation task is carried out to produce the appropriate text to describe the given graph in an auto-regressive manner. We adopt the standard negative log-likelihood loss $\mathcal{L}_{\rm TG}$ for the graph-to-text generation task:
\begin{equation}
\label{equation:teacher-l}
\small
   \mathcal{L}_{\rm TG} = -\frac{1}{N}\sum_{i= 1}^{n} \log p(y_{i}|y_{1},...,y_{i-1};\mathcal{G}_{\rm linear},\mathcal{E})
\end{equation}
where $n$ is the length of the target sequence $Y$.

\section{Experimental Setup}
\subsection{Tasks and Datasets}

To verify the generality and effectiveness of UniD2T, we conduct experiments on three types of data-to-text datasets. In particular, WebNLG \cite{gardent2017creating} and DART \cite{nan2020DART} are used for evaluating graph-to-text generation; WikiBio \cite{lebret2016neural} and WikiTableT \cite{chen2021WikiTableT} are utilized for evaluating key-value-to-text generation; ToTTo \cite{parikh2020ToTTo} and CoSQL \cite{yu-etal-2019-CoSQL} are used for evaluating table-to-text generation.  Table~\ref{tab:test-data} provides the statistics of these six datasets. 

\begin{table}[ht]
\centering
 \small
\begin{tabular}{@{}lrrr@{}}
\toprule
% \hline
\textbf{Dataset}         & \textbf{Train} & \textbf{Valid} & \textbf{Test}  \\ 
               \midrule
ToTTo & 120,761 & 7,700 & 7,700\\
CoSQL & 7,845 & 1,074 & -\\
WebNLG &13,211  & 1,667 & 1,779\\
DART & 62,659 &  6,980 & 12,552 \\
WikiBio  & 582,657 & 72,831 & 72,831\\ 
WikiTableT  & 1,453,794 & 4,533 &  4,351\\ 
\bottomrule
\end{tabular}
\caption{\label{tab:test-data}Statistics of downstream datasets.}
\end{table}

\subsection{Implementation Details}
In the pre-training stage, our model is initialized with T5-Large. We pre-train our UniD2T model on  NVIDIA A100 GPUs. The maximum sequence lengths of the input and target sequences are set to 1024 and 512, respectively. 
We set the batch size to 8. 
Gradient clipping is applied to the model with a maximum gradient value of 1. To alleviate the overfitting issue, the maximum number of training steps is 500k. Moreover, a patient step number is set to 25k, i.e., if the evaluation metrics does not increase for the patient step number, the training process will carry out an early stop. We set the maximum learning rate to 1e-5.  
%We first pre-train the model on UnStruct2Text, and then continue training it on AnStruct2Text.

%We split the downstream tasks into three parts based on the input format, i.e., table-based (ToTTo, CoSQL), graph-based (WebNLG, DART) and key-value-based (WikiBio, WikiTableT, E2ENLG). We conduct few-shot experiments on E2ENLG and fully supervised experiments on other datasets.

% \subsection{Pretrain}
% We unify various data forms in the form of UDS as Method \ref{UDS}, and each dataset is shown in the figure below. We randomly extract 40\% of the data to do the MLM task, that is, cover some of the nodes to predict the covered content, and the remaining 60\% of the data is used for supervision tasks. We set 80 epochs for pre-training, the learning rate is set to 2e-4 and the linear learning rate decay strategy is adopted. Next, we will finetune the obtained pre-trained model downstream and the finetune settings and experimental results are presented in the following subsections.

\begin{table*}[]
\centering
\resizebox{2\columnwidth}{!}{
\begin{tabular}{@{}lcccccc@{}}
\toprule

\multicolumn{1}{c}{\multirow{2}{*}{\textbf{Models}}} & \multicolumn{2}{c}{\textbf{Overall}} & \multicolumn{2}{c}{\textbf{Overlap}} & \multicolumn{2}{c}{\textbf{Non-Overlap}} \\ \cmidrule(r){2-3}  \cmidrule(r){4-5} \cmidrule(r){6-7}
% \cmidrule(l){2-7} 
\multicolumn{1}{c}{}                                 & \textbf{BLEU}    & \textbf{PARENT}   & \textbf{BLEU}    & \textbf{PARENT}   & \textbf{BLEU}      & \textbf{PARENT}     \\ \midrule
ChatGPT(gpt-3.5-turbo)                                            & 20.5             & 49.5              & 24.4             & 51.2              & 17.5               & 47.7               \\

BERT-to-BERT\cite{rothe2020leveraging}                                            & 44.0             & 52.6              & 52.7             & 58.4              & 35.1               & 46.8                \\
LATTICE    \cite{wang2022robust}                                           & 48.4             & 58.1              & 56.1             & 62.4              & 40.4               & 53.9                \\
CoNT  \cite{an2022cont}                                               & 49.1             & 58.9              & 56.7             & 63.2              & 41.3               & 54.6                \\
PlanGen    \cite{su2021plan}                                          & 49.2             & 58.7              & 56.9             & 62.8              & 41.4               & 54.2                \\ \midrule
T5-3B                                              & 49.5             & 58.4              & 57.5             & 62.6              & 41.4               & 54.2                \\
TABT5       \cite{andrejczuk2022table}                                         & 49.2             & 57.2             & -                & -                 & 41.0               & 52.7                \\
\rowcolor[RGB]{237,237,237}
UniD2T                                                 & \textbf{49.9 }            & \textbf{59.8  }            & \textbf{57.8 }            & \textbf{64.0  }            & \textbf{42.0 }              & \textbf{55.7 }               \\ \bottomrule
\end{tabular}
}
\caption{Results on the ToTTo test set. %All reported results, including ours, can be found in the official Leaderboard.\footnote{https://github.com/google-research-datasets/ToTTo}
}
\label{tab:res_ToTTo}

\end{table*}

\begin{table}[ht]
\centering
   \small
\resizebox{0.8\columnwidth}{!}{
\begin{tabular}{lcc}
\toprule

\textbf{Models}         & \textbf{BLEU} & \textbf{ROUGE-L}  \\ \midrule
GraphWriter & 16.86 & 47.44\\
FALCON & 25.65 & 57.89 \\ \midrule[0.2pt]
BART-Base & 24.60 & 57.39\\
T5-Large              & 25.25  & 57.54\\

\rowcolor[RGB]{237,237,237}
UniD2T           & \textbf{32.68} &\textbf{ 61.47}
\\ \bottomrule
\end{tabular}
}
\caption{\label{tab:result-CoSQL}Results on CoSQL development set.}
\end{table}

\begin{table}[ht]
\centering
\resizebox{\columnwidth}{!}{
\begin{tabular}{lccc}

\toprule
                   \textbf{Models}    &\textbf{ BLEU}  & \textbf{METEOR }& \textbf{TER}  \\ \midrule

End-to-End Transformer\dag & 27.24 & 0.25   & 0.65 \\
LSTM with Attention\dag    & 29.66 & 0.27   & 0.63 \\
CONTROL PREFIXES      & 51.95 & 0.41   & 0.43 \\\midrule[0.1pt]
ChatGPT(gpt-3.5-turbo)  & 40.51 & 0.37   & 0.53 \\                               
BART-Base\dag              & 47.11 & 0.38   & 0.46 \\
BART-Large\dag             & 48.56 & 0.39   & 0.45 \\
T5-Small\dag               & 47.69 & 0.39   & 0.46 \\
T5-Base\dag                & 49.21 & 0.40   & 0.44 \\
T5-Large\dag               & 50.66 & 0.40   & 0.43 \\
\rowcolor[RGB]{237,237,237}
UniD2T & {\textbf{54.96}} & {\textbf{0.42}} & \textbf{0.42} \\ \bottomrule
\end{tabular}
}
\caption{\label{result:DART} Evaluation results on DART test set. Results with \dag are token from  DART~\cite{nan2020DART}.}
\end{table}

\begin{table}[]

 \resizebox{\columnwidth}{!}{
\begin{tabular}{lccccc}
\toprule
\textbf{Model}                 & \textbf{BLEU}  & \textbf{METEOR} & \textbf{chrF++} & \textbf{TER}   & \textbf{BLEURT} \\ \midrule

CP       & 54.97 & 41.7  & 69.3  & 39.8  & 0.62   \\
CP + DART & 55.41 & 41.9  & 69.8  & 39.2  & 0.63   \\\midrule[0.1pt]
T5-Large                & 51.74 & 40.3  & 66.9  & 41.7  & 0.61   \\
TRIPLE                  & 57.64 & 42.24  & -      & 38.9 & -      \\
\rowcolor[RGB]{237,237,237}
UniD2T                    & \textbf{60.41} & \textbf{44.35}  &    \textbf{73.4}    & \textbf{34.1}  & \textbf{0.65}   \\ \bottomrule

\end{tabular}
}

\caption{\label{tab:res_WebNLG} Evaluation results on WebNLG test set. CP stands for CONTROL PREFIXES~\cite{clive2021control}.}
\end{table}

\begin{table}[t]
\small
\resizebox{\columnwidth}{!}{
\begin{tabular}{lcccc}
\toprule
 & \multicolumn{2}{c}{\bf WikiBio} & \multicolumn{2}{c}{\bf WikiTableT}\\\cmidrule(r){2-3}  \cmidrule(r){4-5}

\multirow{-3}{*}{\bf Models} & \textbf{BLEU} & \textbf{PARENT} & \textbf{BLEU} & \textbf{PARENT} \\
\midrule
Transformer & 44.3 & 74.0 & 19.5  & 42.8 \\ 
% R2D2            & 46.2 & - & - & -\\
SANA            & 45.7 & 76.9 & - & -\\
CoNT            & 47.1 & - & - & -\\ 
% \multicolumn{5}{c}{\em Finetuning pattern} \\
\midrule[0.1pt]
KGPT & 45.1 & 76.3 & 31.8 & 48.5 \\
% T5-small        & 46.0 & - & - & - \\
T5-Large        & 48.6 & 77.5 & 31.4 & 47.6\\
\rowcolor[RGB]{237,237,237}
UniD2T             & \textbf{50.4} & \textbf{79.8} &\textbf{ 33.7 } &\textbf{50.7}\\ 
\bottomrule
\end{tabular}
}
\caption{Results on WikiBio and WikiTableT test sets.}
\label{tab:WikiBio_and_WikiTableT}
\end{table}

\begin{table*}[ht]
\centering
\resizebox{2\columnwidth}{!}{
\begin{tabular}{@{}lccccccc@{}}
\toprule
     \textbf{Model}             & \textbf{ToTTo}  & \textbf{CoSQL} & \textbf{DART}  & \textbf{WebNLG} & \textbf{WikiBio} & \textbf{WikiTableT} & \textbf{Total Score} \\ \midrule
\multicolumn{8}{c}{\em Only Fine-tuning} \\
\midrule
Previous SOTA            & 49.2 & 25.6 & 51.9 & 57.6       & 48.6   & 31.8 & -        \\ 
T5-Large +$F_{\rm Linear}$             & 48.1 & 25.2  & 50.6 & 51.7   & 48.6       & 31.4 & 255.6      \\ 
T5-Large +$F_{\rm Graph}$            & 49.1   & 26.7  & 51.2 & 53.1 & 49.4   &   32.2  & 261.7       \\ 
\midrule
\multicolumn{8}{c}{\em With Additional Pre-training} \\
\midrule
T5-Large +$P_{\rm Graph}+F_{\rm Graph}$ (UniD2T)         & \textbf{50.2 }&\textbf{32.7}    & \textbf{54.9} & \textbf{60.4}         & \textbf{50.4}   & \textbf{33.7} & \textbf{282.3}    \\ 
T5-Large +$P^{*}_{\rm Graph}+F_{\rm Graph}$ & 49.3  & 27.9  & 53.6 & 54.7         & 50.1   & 32.4 & 268.0    \\ 
T5-Large +$P_{\rm Linear}+F_{\rm Linear}$  & 48.7   & 25.8 & 53.1    & 56.7 &     49.1    &     31.7 &   265.1   \\
\griit{T5-Large} +$\griit{P^{*}_{\rm Linear}+F_{\rm Linear}}$ & 48.3  & 25.7  & 50.9 & 52.8         & 48.7   & 31.5 & 257.9   
\\\bottomrule
\end{tabular}
`}
\caption{ Ablation test results on six benchmark datasets. $P_{\rm Linear}$ and $P_{\rm Graph}$ represent the models pre-training with linear structure and graph structure, respectively.  $F_{\rm Linear}$ and $F_{\rm Graph}$ represent the models fine-tuning with graph structure and linear structure, respectively. $P^{*}$ stands for pre-training only with \textsc{PreData}; $P$ indicates pre-training with both \textsc{PreData} and \textsc{DownData}.}
\label{tab:ablation_res}
\end{table*}

\begin{table}[]
% \small
\resizebox{\columnwidth}{!}{
\begin{tabular}{@{}lccccc@{}}
\toprule
{\color[HTML]{262626} \textbf{Models}} & \multicolumn{1}{l}{\textbf{BLEU}} & \multicolumn{1}{l}{\textbf{METEOR}} & \multicolumn{1}{l}{\textbf{chrF++}} & \multicolumn{1}{l}{\textbf{TER}} & \multicolumn{1}{l}{\textbf{BLEURT}} \\ \midrule
{\color[HTML]{262626} UniD2T} & \textbf{60.4}  &\textbf{ 44.4} & \textbf{73.4} &\textbf{ 34.1} & \textbf{0.65} \\
\quad - attention                  & 58.6  & 42.7 & 70.3 & 37.2 & 0.64 \\
\quad - position                   & 58.3  & 42.6 & 70.2 & 36.7 & 0.64 \\
\quad - all                        & 56.7  & 42.3 & 69.8 & 37.8 & 0.63 \\ \bottomrule
\end{tabular}
}
\caption{\label{tab:ablation2} Ablation test results on WebNLG test set.}
\end{table}

% \subsection{Effectiveness of pre-training}
% \subsection{Effectiveness of graph structure}

\section{Experimental Results}

\subsection{Table-to-Text Generation}
We conduct experiments on two table-to-text datasets, including ToTTo and CoSQL. \griit{The SQL queries within CoSQL and the table header information from ToTTo are strategically positioned within the data-specific prefixes, denoted as ``[\textit{Prefix-S}]'', as illustrated in Table~\ref{tab:prefix}.}

% The experimental results on the two datasets are summarized in Table~\ref{result:ToTTo} and Table~\ref{tab:result-CoSQL}, respectively. Our model achieves substantially better performance than the compared competitors on ToTTo development in terms of overall, overlap, and non-overlap settings. The results show the advantages of our model in generating text with fluency and faithfulness. \dll{CoSQL is a response generation dataset that requires generating a natural language description of the SQL and the execution result table. Our model achieves SOTA results on it. The results indicate}

\paragraph{ToTTo} 
ToTTo is an open-domain table-to-text task dataset that uses crowd annotators to highlight the table cells and revise the corresponding natural language descriptions.
%\Grit{\delete{We finetune our model on ToTTo for 20 epochs. The learning rate is set to 3e-5, the maximum input length is limited to 1024, the maximum target length is 208, and the beam search is set to 5.}}
We compare our UniD2T with several strong baselines, including BERT2BERT \cite{rothe2020leveraging}, LATTICE \cite{wang2022robust}, CoNT \cite{an2022cont}, PlanGen \cite{su2021plan} and TABT5~\cite{andrejczuk2022table}. TABT5 is a pre-trained model tailored for table-to-text generation. We adopt BLEU \cite{papineni2002BLEU} and PARENT \cite{dhingra2019handling} as the evaluation metrics. 
%\begin{itemize}
%\item BERT-to-BERT \cite{rothe2020leveraging}: a Transformer-based sequence-to-sequence model that uses pre-trained bert checkpoint to initialize and share the parameters at both ends of the encoder and decoder.
%\item LATTICE \cite{wang2022robust}: an equivariance learning framework using structure-aware self-attention to encode Tables.
%\item CoNT \cite{an2022cont}: a new framework for contrastive neural text generation that improves on contrastive learning in generative tasks from contrastive examples, losses and decoding strategies.
%\item PlanGen \cite{su2021plan}: a plan-to-generate framework to control the generated intra-sentential and inter-sentential structures.
%\end{itemize}
%Among these baselines, some are designed for specific input format, such as LATTICE for table-to-text generation tasks, and some are for broad generation tasks, whose input is often serialized. 
The experimental results on ToTTo are summarized in Table~\ref{tab:res_ToTTo}. Our model achieves substantially better performance than the compared methods on ToTTo in terms of overall, overlap, and non-overlap settings. First, our model \griit{shows an improvement over T5 and TABT5, especially in terms of PARENT}. Second, our model also achieves better results than the strong downstream methods. 
%For example, compared with the pre-trained model T5-3B, even though it has 3x fewer parameters.
%\grit{Compared with the pre-trained model, UDon outperforms T5-3B, even though it has 3x fewer parameters. Compared with the structure planGen designed specifically for the table, because pre-training can have gains from other data sets, in non-overlap, it improves 0.6 BLEU and 1.5 PARENT, but our more improvement is reflected in the overlap table, with 0.9 BLEU The improvement and the parent improvement of 1.2 prove that the graph structure we proposed is also necessary and effective.}
%Our model achieves SOTA results in all three dimensions (Overall, Overlap, Non-Overlap) and two evaluation indicators: BLEU and PARENT. The Non-Overlap set features examples that are out-of-domain from the training set.

% \usepackage{booktabs}

\paragraph{CoSQL}
\griit{CoSQL serves as a prevalent benchmark for evaluating table-to-text models~\citep{fang2022falcon, li2023cats}. Each instance within CoSQL comprises an SQL query, the resultant table, and the corresponding response, where the SQL query gives explicit signals for models on what to generate. The generated description could provide a concise and easy-to-understand summary of the result table and help users verify whether the queried result is consistent with the original question.}
%CoSQL is a response generation dataset that servers for the practical TableQA system. The dataset requires generating a natural language description of the SQL and the execution table. 
% Specifically, the SQL is generated by a semantic parsing module, which converts the users' natural language questions into SQL queries. CoSQL is pragmatic and challenging. It requires the model to understand two heterogeneous source data, a SQL with complex grammar and a structural table, and then provide a concise and easy-to-understand summary about the result table. 
We compare our model with GraphWriter~\cite{koncel2019text}, BART-Base, T5-Large, and FALCON~\citep{fang2022faithful} that is a faithful contrastive generation framework based on T5. We adopt BLEU \cite{papineni2002BLEU} and ROUGE-L \cite{lin2004rouge} as evaluation metrics. 
Since CoSQL does not release the test set, we follow FALCON and report the experimental results on the development set in Table~\ref{tab:result-CoSQL}.
%\delete{ Our model achieves significantly better performance than T5-Large and FALCON. }
Our UniD2T model achieves significantly better performance than baselines. The BLEU and ROUGE scores increase by 7.03 and 3.58 respectively over the best-performing baseline FALCON. 
%UniD2T improves upon the T5-Large basline by  7.43 BLEU and 3.93 ROUGE. Moreover, Compared with the baseline FALCONb, which considers SQL characteristics for CoSQL, we have improved by 7.43 BLEU and 3.58 ROUGE.

\subsection{Graph-to-Text Generation}
We conduct experiments on two graph-to-text datasets, including DART and WebNLG. 
%\delete{Following \cite{nan2020DART} for both datasets, the learning rates are set to 7e-5; the maximum input and output length is set to 384.}

\paragraph{DART}
DART is a large dataset for open-domain text generation that treats the input as a set of RDF entity-relation triples. 
%Connections between triples form graph struct.
%\delete{We fine-tune pre-trained models on this dataset.}
We compare our UniD2T model with several pre-training models including Transformer, BART, T5, and the state-of-the-art method CONTROL PREFIXES~\cite{clive2021control}. BLEU \cite{papineni2002BLEU}, METEOR \cite{banerjee2005meteor}, and TER \cite{snover2005study_ter} are adopted as evaluation metrics. 
As shown in Table~\ref{result:DART}, our model 
surpasses the best-performing model CONTROL PREFIXES by a 3.0\% BLEU. 
%\grit{Moreover, higher METEOR and lower TER also reflect that the sentences generated by the model are more fluent and closer to ground truth.}
%outperforms CONTROL PREFIXES by more than 3 points on BLEU and also improves in METEOR and TER evaluation indicators.

\paragraph{WebNLG} 
WebNLG \cite{zhou-lampouras-2020-WebNLG} consists of a set of triples collected from DBpedia and the corresponding manually annotated text. 
%There are many versions of WebNLG, we use the latest version WebNLG+2020 \cite{zhou-lampouras-2020-WebNLG}, which has 16 distinct DBpedia categories.
%WebNLG consists of a sequence of triples and the text that verbalization of these triples. 
%There are many versions of the WebNLG dataset. We use the 2017 version like most other baseline settings. 
BLEU \cite{papineni2002BLEU}, METEOR \cite{banerjee2005meteor},  chrF++ \cite{popovic2015chrf}, TER~\cite{snover2005study_ter} and BLEURT~\cite{sellam2020bleurt} are adopted as evaluation metrics. 
%the training set has 10 visible categories, but other invisible 5 categories appear on the test set, so the results are divided into three parts: overall, seen, and unseen. 
We compare our method with both pre-trained language models and strong downstream baselines.
The overall experimental results on WebNLG are shown in Table~\ref{tab:res_WebNLG}. Our model achieves the highest performance among all baseline models, including the graph pre-training model TRIPLE \cite{han2022self_tripple}.

% \usepackage{booktabs}

\subsection{Key-Value-to-Text Generation}
We conduct experiments on two key-value-based datasets, including WikiBio and WikiTableT. 

%\yangmin{Pls modify this section following Section 6.1 and 6.2. Include evaluation metrics, baselines, results, etc.}

\paragraph{WikiBio} WikiBio is designed to generate descriptions from a Wikipedia infobox and aims to generate the first sentence of a biography. We compare UniD2T with previous state-of-the-art model (i.e, CoNT~\cite{an2022cont}), pre-trained models (T5-Large, KGPT) and Non-autoregressive model SANA~\cite{sana_WangLYZZZY21} on WikiBio. BLEU  and PARENT are adopted as evaluation metrics. The results are reported in Table \ref{tab:WikiBio_and_WikiTableT}. UniD2T outperforms the best baseline CoNT by 3.7\% on BLEU.

\paragraph{WikiTableT} WikiTableT is collected from Wikipedia sections with their corresponding tabular data, which contains millions of instances. We compare UniD2T with Transformer, T5-Large and KGPT~\cite{chen2020kgpt}.
Experiment results on Table~\ref{tab:WikiBio_and_WikiTableT} show that UniD2T exceeds the best competitor KGPT by 1.9\% on BLEU and 2.2\% on PARENT. 

%achieves new state-of-the-art results on both of the Key-Value-based datasets, evaluated by BLEU~\cite{papineni2002BLEU} and PARENT~\cite{dhingra2019handling}.

%\yangmin{For myself, modify from here.}

%In summary, the proposed UniD2T achieves SOTA results on three types of data-to-text tasks without any special design dedicated to different data types. This demonstrates the effectiveness of our unified data-to-text pretraining method.

\subsection{Further Analysis}
\subsubsection{Ablation Study} 
%\paragraph{Impact of Pre-training with Graph Structure and Linear Structure}
We conduct experiments to investigate the impact of pre-training with graph structure and linear structure. The ablation results are summarized in Table~\ref{tab:ablation_res}, which is divided into two parts: the first part shows the results of directly fine-tuning the pre-trained language model  (i.e., T5-Large) on the downstream datasets, referred to as \textsc{DownData}, while the second part presents the results of incorporating additional pre-training data, denoted as \textsc{PreData}, on top of T5-Large. Through careful analysis, we observe that UniD2T (T5-Large+$P_{\rm Graph}$+$F_{\rm Graph}$) consistently outperforms T5-Large+$F_{\rm Graph}$ across all six data-to-text datasets, resulting in a notable improvement in the total score of \textbf{$+20.6$}. %In addition, while T5-Large+$F_{\rm Linear}$ achieves a total score of $255.6$ on the six data-to-text datasets,  T5-Large+$P_{\rm Linear}+F_{\rm Linear}$ enhances the total scores by \textbf{$+9.5$}. %These findings clearly indicate that continual pre-training significantly enhances the performance of previous pre-training language models on data-to-text tasks.
%Furthermore, we delve into the impact of our unified structured data modeling, which unifies various data formats into a graph structure and explicitly encodes its underlying structure. 
In addition, we observe that T5-Large+$F_{\rm Graph}$ outperforms T5-Large+$F_{\rm Linear}$ in terms of the total score by $+6.1$. This result clearly indicates that our method significantly improves the performance of the data-to-text models which linearize the structured data as input during fine-tuning the models on downstream datasets. %Similarly, by comparing the results of UniD2T and T5-Large+$P_{\rm Linear}+F_{\rm Linear}$, our method enhances the performance of the models that ignore the graph structure. 
Finally, we delve into the effects of the pre-training datasets. By comparing the results of $P^{*}_{\rm Graph}+F_{\rm Graph}$ and $P_{\rm Graph}+F_{\rm Graph}$, $\griit{P^{*}_{\rm Linear}+F_{\rm Linear}}$ and $\griit{P_{\rm Linear}+F_{\rm Linear}}$, we observe that the downstream datasets contribute to improving the model's performance and accelerating the pre-training process. It is noteworthy that the pre-training involving both \textsc{PreData} and \textsc{DownData} achieves the best performance across all the experimental datasets.

We also delve into the effects of two Transformer modifications (position and attention matrix construction). The results are illustrated in Table~\ref{tab:ablation2}. From the results, we observe a significant performance drop when either the structure-aware position or attention matrices are removed, demonstrating the benefits of two Transformer modifications. It is no surprise that combining all the factors achieves the best performance. These findings collectively demonstrate the effectiveness of our proposed method, which explicitly models its graph structure through the use of structure-aware position and attention matrices.

\begin{table}[ht]
\centering
%\resizebox{0.5\textwidth}{!}{
    % \small
\begin{tabular}{lcccc}
\toprule
% \rowcolors{1}{blue!20}{blue!10}
\textbf{Model}            & \textbf{0.1\%} & \textbf{0.5\%} & \textbf{1\%} & \textbf{5\%} \\
\midrule
TGen              & 3.6 & 27.9 & 35.2 & 57.3     \\
Template-GPT-2    & 22.5 & 47.8 & 53.3 & 59.9 \\
KGPT-Graph       & 39.8 & 53.3 & 55.1 & 61.5 \\
KGPT-Seq      & 40.2 & 53.0 & 54.1 & 61.1 \\
UniD2T            & \textbf{45.6} & \textbf{57.3 }& \textbf{57.6} & \textbf{64.8} \\ 
\bottomrule
\end{tabular}
%}
\caption{\label{tab:few-shot}\griit{Few-shot results on the E2ENLG test set.}%\grit{with 0.1\%, 0.5\%, 1\%, 5\%, 10\% training instences.}
}
\end{table}

\subsubsection{Few-Shot Results}
\griit{We conduct few-shot experiments on the E2ENLG \cite{dusek.etal2020:csl}  dataset sourced from the restaurant domain other than Wikipedia. This serves as an additional validation of the model's generalization capabilities. The E2ENLG dataset, assembled through the CrowdFlower platform, encompasses details about restaurants and comprises over 50,000 combinations of dialogue-act-based meaning representations (MR) with an average of 8.1 references. We fine-tune UniD2T using varying proportions of the training instances (i.e., 0.1\%, 0.5\%, 1\%, 5\%, and 10\%) from E2ENLG \cite{dusek.etal2020:csl}. We compare UniD2T with several few-shot learning methods including TGen \cite{duvsek2016sequence}, Template-GPT-2 \cite{chen2020logical}, and KGPT \cite{chen2020kgpt}.  The experimental results are summarized in Table \ref{tab:few-shot}. We can see that UniD2T significantly outperforms all baselines in various few-shot settings.}

%The few-shot learning setting aims to investigate the generality of UniD2T. \griit{In this study, we fine-tuned our model using varying proportions of the training instances, specifically 0.1\%, 0.5\%, 1\%, 5\%, and 10\% from the E2ENLG \cite{dusek.etal2020:csl} dataset. E2ENLG is a specialized dataset designed for the task of transforming structured restaurant data into coherent natural language descriptions. This dataset encompasses features such as cuisine type, price range, and ratings, paired with their respective textual representations.} 
%The few-shot scenario may pose unseen entities during testing time. We compare UniD2T with several few-shot learning methods including TGen \cite{duvsek2016sequence}, Template-GPT-2 \cite{chen2020logical}, KGPT \cite{chen2020kgpt}. The few-shot results are summarized in Table \ref{tab:few-shot}. We can see that UniD2T significantly outperforms all baselines in various few-shot settings. 

\begin{figure}[t!]
\centering
\includegraphics[width=1.0\columnwidth]{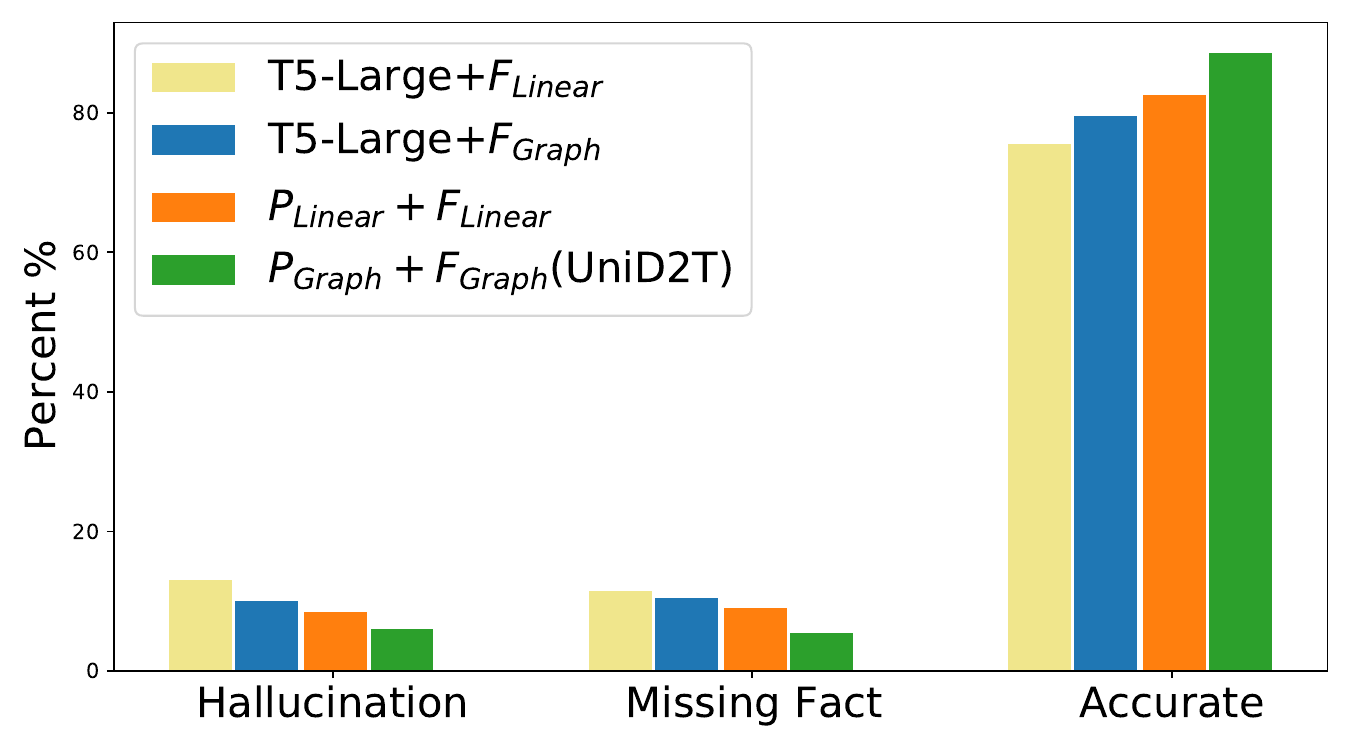}
\caption{Human evaluation of the factual consistency of different models on WebNLG samples.}
\label{fig:human_eval}
\end{figure}
    
\subsubsection{Human Evaluation}
We also conduct a human evaluation to analyze the generated sentences following \citet{chen2020kgpt}. It is worth noting that each evaluator is unaware of which model generates the text being evaluated so as to avoid evaluation bias. Specifically, we choose 100 test samples from WebNLG and observe the factual consistency between the gold sentences and generated sentences. We invite four NLP workers to assign each text a label from \{\textit{Hallucination}, \textit{Missing Fact}, \textit{Accurate}\}, similar to \cite{chen2020kgpt}. As shown in Figure~\ref{fig:human_eval}, our UniD2T is less prone to hallucinating non-existing facts and can generate more accurate sentences. 

    \begin{figure}[t!]
    \centering
    \includegraphics[width = 7.5cm]{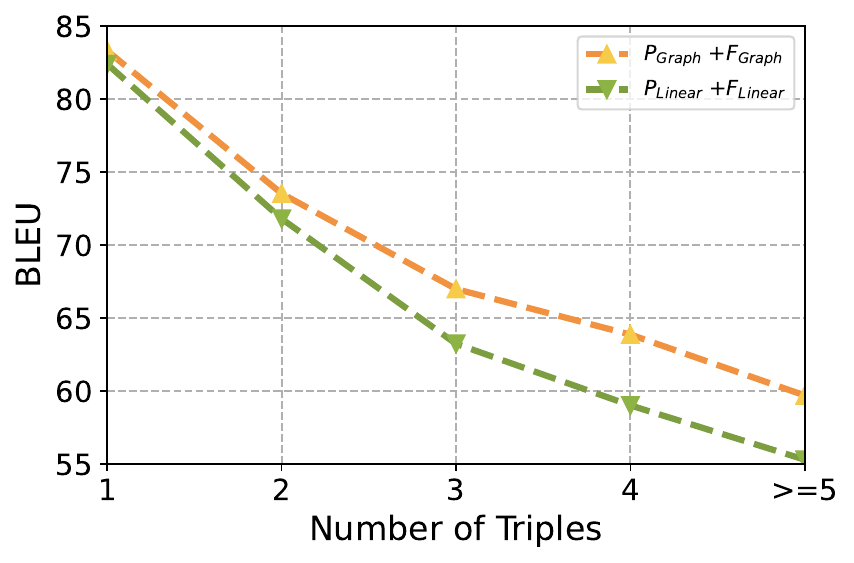}
    \caption{\label{fig:graph_size_1}Comparing $P_{Linear} + F_{Linear}$ and $P_{Graph} + F_{Graph}$ BLEU score changes in increasing the number of triples on WebNLG's seen and unseen.}  
    \end{figure}
    
\subsubsection{Impact on Graph Sizes} 
To illustrate the effectiveness of the graph structure, we further investigate the performances of $P_{\rm Linear}+F_{\rm Linear}$ and $P^{*}_{\rm Graph}+F_{\rm Graph}$ by concerning different graph sizes on the WebNLG validation set.  Experimental results in terms of BLEU are shown in Figure~\ref{fig:graph_size_1}. When the graph structure is simple, the impact of the graph structure is limited. However, as the graph structure becomes complex, the model with graph structure ($P^{*}_{\rm Graph}+F_{\rm Graph}$) performs much better than the model with linear structure ($P_{\rm Linear}+F_{\rm Linear}$). Thus, the structure-enhanced model UniD2T demonstrates greater stability and better performance on large-scale inputs when compared to linear sequence models.

\subsubsection{Impact on Model Sizes}
To investigate the influence of different model scales on the experimental results, we conducted experiments using $F_{\rm Graph}$ on \griit{T5-Small, T5-Base, T5-Large, and T5-3B on the DART and ToTTo dev sets without pre-training}. It is important to note that for our experiments, we conduct evaluations on the dev sets rather than the test sets. This decision is made due to the constraints imposed by the ToTTo dataset, where obtaining test results requires submitting predictions to the leaderboard and awaiting the evaluation process, which can be time-consuming. Therefore, to expedite our research and streamline the experimentation process, we relied on the readily available development sets for conducting our evaluations. The results are presented in the Table \ref{tab:model}. Notably, the transition from T5-Large to T5-3B resulted in a substantial increase in the number of parameters by approximately 3.9 times. However, the corresponding improvement in efficacy was found to be less than 1\%. This analysis sheds light on the limited impact of scaling up the model size beyond a certain threshold, given the marginal gains in performance despite the significant increase in parameter count.

%\grit{In order to demonstrate the impact of different model scales on experimental results, yet constrained by resource availability and time, we executed experiments using T5-small, T5-base, T5-Large, and T5-3B on the DART and ToTTo dev set without pre-trainin, respectively. The results are presented in the Table \ref{tab:model}.Undoubtedly, an increase in scale engenders an enhancement in performance. However, the transition from T5-Large to T5-3B witnesses a multiplication in the number of parameters by approximately 3.9 times, while the corresponding augmentation in efficacy does not surpass 1\%.}
\begin{table}[]
\centering
\resizebox{\columnwidth}{!}{
% \usepackage{booktabs}

\begin{tabular}{@{}lccccc@{}}
\toprule
         & \multicolumn{2}{c}{\textbf{ToTTo}} & \multicolumn{3}{c}{\textbf{DART}} \\ \cmidrule(r){2-3}  \cmidrule(r){4-6}
         & \textbf{BLEU}       & \textbf{PARENT}       & \textbf{BLEU}   & \textbf{METEOR}  & \textbf{TER}   \\ \midrule
\griit{T5-Small} +$F_{\rm Graph}$ & 45.5       & 53.3         & 48.8   & 0.39    & 0.45  \\
\griit{T5-Base} +$F_{\rm Graph}$   & 48.6       & 58.8         & 50.2   & 0.40    & 0.44  \\
\griit{T5-Large} +$F_{\rm Graph}$  & 49.1       & 59.4         & 51.2   & 0.40    & 0.43  \\
\griit{T5-3B} +$F_{\rm Graph}$    & 49.8       & 59.7         & 51.4   & 0.41    & 0.43  \\ \bottomrule
\end{tabular}
}
\caption{\label{tab:model} 
The performance of T5 with different model scales on the dev sets of DART and ToTTo datasets, without performing any pre-training. 
}
\end{table}

\iffalse
\begin{table}[]
\begin{tabular}{@{}ll@{}}
\toprule
\multicolumn{2}{c}{\textbf{ToTTo}} \\ \midrule
Prompt       &   \makecell[l]{Put the highlighted-table  together to \\ form a sentence: }     \\ \midrule[0.1pt]
Structured Input     &    \makecell[l]{ <page\_title> List of Malayalam films\\ of 1976 </page\_title><table> <cell>\\ Surveykkallu <col\_header> Film\\ </col\_header> {}</cell> {} <cell>\\ Thoppil Bhasi <col\_header> Director\\ </col\_header> </cell> </table>     }      \\ \midrule[0.8pt]
\multicolumn{2}{c}{\textbf{DART}}  \\ \midrule
Prompt        &   \makecell[l]{Put the triples together  to form a \\ sentence: }      \\ \midrule[0.1pt]
Structured Input    &   \makecell[l]{Mars Hill College : joined : 1973 |\\ Mars Hill College : location : Mars \\ Hill, North Carolina}  \\ \bottomrule
\end{tabular}
\caption{\label{tab:ChatGPT_input} Input example using ChatGPT on ToTTo and DART, prompt represents task description, and struct represents specific data input.}
\end{table}
\fi

\begin{table}[]
\small
\begin{tabular}{p{7cm}}
\toprule
\multicolumn{1}{c}{\textbf{ToTTo}}\\ \midrule
\textsc{Prompt}: Put the highlighted-table  together to form a sentence:  \\ \midrule[0.1pt]
\textsc{Structured Input}: <page\_title> List of Malayalam films of 1976 </page\_title><table> <cell> Surveykkallu <col\_header> Film </col\_header> {}</cell> {} <cell> Thoppil Bhasi <col\_header> Director </col\_header> </cell> </table>        \\ \midrule[0.8pt]
\multicolumn{1}{c}{\textbf{DART}}  \\ \midrule
\textsc{Prompt}: Put the triples together  to form a  sentence:       \\ \midrule[0.1pt]
\textsc{Structured Input}: Mars Hill College : joined : 1973 | Mars Hill College : location : Mars Hill, North Carolina \\ \bottomrule
\end{tabular}
\caption{\label{tab:ChatGPT_input} Input examples for ChatGPT on ToTTo and DART. Here, \textsc{Prompt} represents task description, and \textsc{Structured Input} represents data input with specific formats.}
\end{table}

\subsection{The Zero-shot Performance of ChatGPT}
We conducted zero-shot experiments using ChatGPT on the ToTTo and DART datasets to establish baselines for performance evaluation. The results of these experiments are presented in Table \ref{tab:res_ToTTo} and Table \ref{result:DART} as baselines. The prompt structure of ChatGPT comprises two parts, and detailed information regarding these prompts can be found in Table~\ref{tab:ChatGPT_input}.

From the results, we observe that ChatGPT demonstrates consistent performance across various measures. For instance, in the non-overlap subset of the ToTTo dataset, when compared to BERT-to-BERT, the BLEU score shows a decrease of 17.6\%, while the PARENT score exhibits a slight increase of 0.9\%. This divergence in BLEU performance indicates that ChatGPT generates responses with different word choices, leading to reduced word overlap with the reference. However, the improvement in the PARENT score suggests enhanced structural and content-related aspects in the generated responses. These findings underscore the importance of employing multiple evaluation metrics to comprehensively assess the performance of sophisticated language generation systems in future work.

\begin{table}[]
% \small
\resizebox{\columnwidth}{!}{
\begin{tabular}{@{}lcccc@{}}
\toprule
\textbf{Models} & \textbf{Distinct-1} & \textbf{Distinct-2} & \textbf{Distinct-3} & \textbf{Distinct-4} \\ \midrule
%Ground-truth & - & - & - & - \\
%\midrule
ChatGPT & \bf 7.56  & \bf 18.93 & \bf 28.33 & \bf 35.75 \\
T5-Large & 6.94 & 13.94 & 19.00 & 23.00 \\
%\midrule
UniD2T & 6.58 & 14.72 & 21.22 & 26.38 \\
\bottomrule
\end{tabular}
}
\caption{\label{tab:diversity} The results of diversity evaluation  on DART test set.}
\end{table}

\begin{table}[]
\resizebox{\columnwidth}{!}{
\begin{tabular}{lcccc}
\toprule
Edge               & \multicolumn{2}{c}{WikiBio} & \multicolumn{2}{c}{WikiTableT} \\ \cmidrule{2-5} 
                    & BLEU        & PARENT        & BLEU          & PARENT         \\ \midrule
UniD2T              & 50.4        & 79.8          & 33.7          & 50.7           \\
UniD2T$_\text{directed}$ & 48.8        & 78.5          & 31.7          & 48.3           \\ 
\bottomrule
\end{tabular}
}
\caption{\label{tab:direction} \griit{The results of our models with undirected graphs (i.e., UniD2T) and directed graphs (denoted as UniD2T$_\text{directed}$), respectively.}}
%Comparison Results between Undirected Edges and Fully Unidirectional Edges on Wikibio and WikitableT. A and B represent nodes in the graph, $\leftrightarrow$ denotes a bidirectional edge between two nodes, and $\rightarrow$  signifies that all nodes are unidirectional, following the input order from left to right.}
\end{table}

\begin{figure*}[ht]
\centering
    \includegraphics[width=1\textwidth]{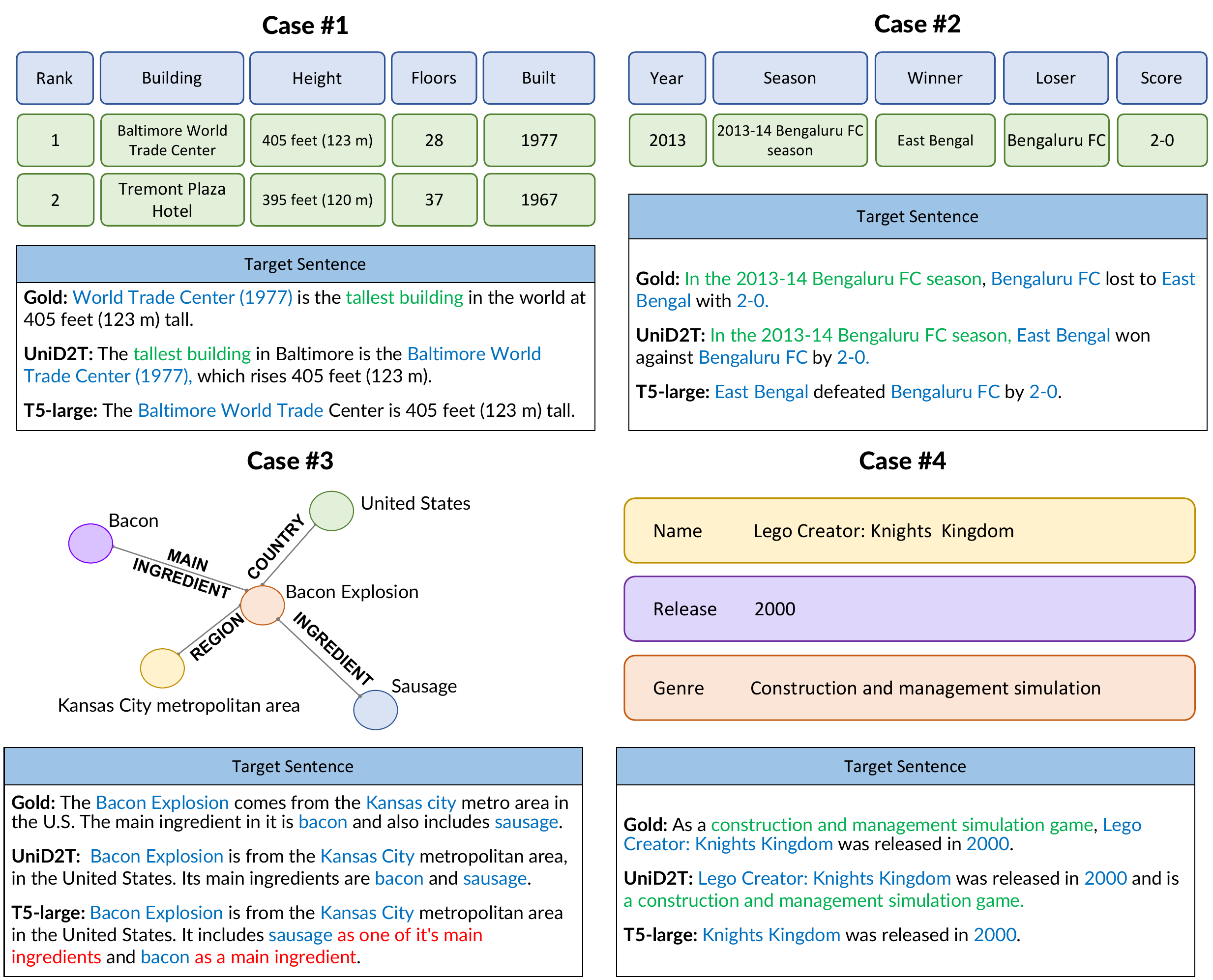}
    \caption{Examples of generated sentences. The main entity is highlighted in green, and the words that are not faithful to the input are in red. Important information common to both models is indicated in blue.}
    \label{fig:case_study}
\end{figure*}

\subsubsection{Impact on Edge Directionality}
\griit{We take an examination into the significance of edge directionality and present the experimental results of incorporating the edge direction in Table \ref{tab:direction}. For UniD2T$_\text{directed}$, we consider the input directed graph using only its original directed edges (uni-directional) and remove the reverse edges added by UniD2T. Please refer to Section~\ref{sec:unifying_structured_data} for more details about the reverse edges.
From Table \ref{tab:direction}, we can observe that the incorporation of edge direction has a deleterious effect on the performance of pre-trained models. There are several possible factors that may underlie these observed outcomes. (1) First, the pre-training models aim to learn the general representations of structured data. However, due to the vast scale of multi-source data, it is often unfeasible to assign a direction to each data pair. For example, the tabular format constitutes a fundamental type of structured data; however, the absence of explicit edge directionality is a typical characteristic between individual data pairs within this format. Therefore, we default to using bidirectional edges to signify mutual relationships between two entities. (2) Second, we anticipate learning the coarse relationships between two entities through undirected graphs during the pre-training phase offer greater flexibility to accommodate various types of relationships in different fields.
% and applications without the need for specific adjustments or customization.
% , while the fine-grained relationships can be effectively captured by fine-tuning the pre-trained models on downstream tasks, employing directed graphs.
For instance, the directional link ``\textit{Jay Chou $\rightarrow$ Common Jasmine Orange}'' conveys that \textit{Jay Chou} released the album \textit{Common Jasmine Orange}, while the reverse link ``\textit{Common Jasmine Orange $\rightarrow$ Jay Chou}'' signifies that \textit{Common Jasmine Orange} is one of \textit{Jay Chou}'s albums. In most cases, it is unnecessary to provide elaborate descriptions of specific relationships, as the data primarily requires indicating connections. }

\subsection{Case Study}
As illustrated in Figure \ref{fig:case_study}, we further verify the effectiveness of UniD2T qualitatively by demonstrating some generated sentences by UniD2T and T5-Large. Both UniD2T and T5-Large are capable of generating main entities. However, there are notable differences in the quality and coherence of the generated sentences. Specifically, the sentences generated by \griit{T5-Large} tend to exhibit shortcomings in terms of including key information and logical reasoning. For instance, in the first case, T5-Large fails to infer that the ``Baltimore World Trade Center'' is the tallest building. This illustrates the limitation of T5-Large in capturing and incorporating specific facts with logical reasoning. In contrast, UniD2T can produce sentences that are more accurate, complete, and encompass the main entities and logical information with greater precision. This highlights the advantages of UniD2T in generating more contextually appropriate and logically grounded sentences.

\subsection{The Diversity of Generated Sentences}
We conduct an evaluation of the diversity exhibited in the target sentences generated by UniD2T  and compare it with strong baselines (i.e., T5-Large and ChatGPT). To quantify the diversity of the generated sentences, we employed the Distinct-N metric~\cite{distinct_metric}, which calculates the number of distinct N-grams divided by the total number of generated tokens. The experimental results are presented in Table ~\ref{tab:diversity}, providing insights into the diversity performance of the models. By analyzing the results, it is evident that UniD2T achieves notably higher Distinct-2/3/4  scores compared to T5-Large. This suggests that UniD2T generates sentences with a greater variety of unique unigrams and bigrams than T5-Large, indicating a higher level of linguistic diversity in the output. However, ChatGPT achieves better diversity scores than UniD2T. It tends to generate more diverse words which are not included in our vocabulary, although these words may be non-existing content.
%\liang{We utilize Distinct N-gram~\cite{distinct_metric} to evaluate the diversity of the output text. The experimental results are summarized in Table~\ref{tab:diversity}. Compared to the similarly-sized T5-Large model, UniDT2 improves the diversity of the generated two-, three-, and four-grams. However, it is worth noting that, despite the poor quality of ChatGPT generation on the DART test set, as demonstrated in Table~\ref{result:DART}, it exhibits a considerably higher degree of text diversity compared to the other models.}

\subsection{Limitations}
%We analyze the limitations of this work, so as to further improve the performance of our model in the future. 
Based on our empirical observation, we reveal several limitations of this work, which can be divided into three primary categories. (1) Our pre-training data is limited, which only contains two existing pre-training datasets and six downstream datasets. In the future, we would like to collect  more data-to-text datasets so as to construct a large-scale diverse pre-training corpus. (2) In this work, we unify different structured data into the graph format by using a simple and direct way. We will attempt to exploit more advanced strategies to construct graphs from different structured data. (3) This study focuses on modeling the graph structures and incorporating the structural information into Transformer. However, the pre-training objectives can be further improved so as to further improve the representation learning. %In future work, we plan to design more advanced pre-training objectives such as contrastive learning loss to learn better data and text representations.  

\section{Conclusion}
In this paper, we proposed a unified data-to-text pre-training method, which could be applied to various downstream data-to-text generation tasks. Concretely, we first converted different types of structured data into graph format. Then, we devised a structure-enhanced Transformer to capture graph structures by introducing two new position and attention matrices to replace the position embedding and attention mask in the self-attention of the Transformer. Extensive experiments on six data-to-text benchmark datasets demonstrated that UniD2T achieved substantially better performance than strong baselines by enabling better information sharing and representation learning of data structures across diverse data-to-text datasets.
%Extensive experiments provide empirical evidence that UniD2T enhances the effectiveness of data-to-text generation by enabling better information sharing and improved representation of data structures across diverse datasets. Our results suggest that UniD2T can be a promising approach for advancing the state-of-the-art in this field.

\section*{Acknowledgments}
Min Yang was supported by National Key Research and Development Program of China (2022YFF0902100), National Natural Science Foundation of China (62376262), Shenzhen Science and Technology Innovation Program (KQTD20190929172835662), Shenzhen Basic Research Foundation (JCYJ20210324115614039 and JCYJ20200109113441941). This work was supported by Alibaba Group through Alibaba Innovative Research Program.

\bibliography{tacl}
\bibliographystyle{acl_natbib}
\clearpage
% \section{\grit{Appendix}}

\end{document}